\documentclass[10pt,twocolumn,letterpaper]{article}

\usepackage{iccv}
\usepackage{times}
\usepackage{epsfig}
\usepackage{graphicx}
\usepackage{amsmath}
\usepackage{amssymb}
\usepackage{booktabs}
\usepackage{algorithmicx}
\usepackage{algpseudocode}
\usepackage{algorithm}
\usepackage{multirow}
\usepackage{colortbl}
\usepackage{subfig}
\usepackage{threeparttable}
\usepackage{bbm}
\usepackage{varwidth}
\usepackage[dvipsnames]{xcolor}

\usepackage[breaklinks=true,bookmarks=false]{hyperref}

\iccvfinalcopy 

\def\iccvPaperID{9898} 

\ificcvfinal\fi

\begin{document}


\title{Few-Shot and Continual Learning with Attentive Independent Mechanisms}

\author{Eugene Lee \quad Cheng-Han Huang \quad Chen-Yi Lee \\
Institute of Electronics, National Chiao Tung University\\
Hsinchu, Taiwan\\
{\tt\small \{eugenelet.ee06g, huang50213.ee04\}@nctu.edu.tw \quad cylee@si2lab.org}
}

\maketitle

\begin{abstract}
   Deep neural networks (DNNs) are known to perform well when deployed to test distributions that shares high similarity with the training distribution. Feeding DNNs with new data sequentially that were unseen in the training distribution has two major challenges --- fast adaptation to new tasks and catastrophic forgetting of old tasks. Such difficulties paved way for the on-going research on few-shot learning and continual learning. To tackle these problems, we introduce Attentive Independent Mechanisms (AIM). We incorporate the idea of learning using fast and slow weights in conjunction with the decoupling of the feature extraction and higher-order conceptual learning of a DNN. AIM is designed for higher-order conceptual learning, modeled by a mixture of experts that compete to learn independent concepts to solve a new task. AIM is a modular component that can be inserted into existing deep learning frameworks. We demonstrate its capability for few-shot learning by adding it to SIB and trained on MiniImageNet and CIFAR-FS, showing significant improvement. AIM is also applied to ANML and OML trained on Omniglot, CIFAR-100 and MiniImageNet to demonstrate its capability in continual learning. Code made publicly available at \url{https://github.com/huang50213/AIM-Fewshot-Continual}.
\end{abstract}

\section{Introduction}
Humans have the ability to learn new concepts continually while retaining previously learned concepts \cite{fagot2006evidence}. While learning new concepts, prior concepts that were learned are leveraged to form new connections in the brain \cite{bauer2015monitoring,zeithamova2019brain}. The plasticity of the human brain plays an important role on the forming of novel neuronal connections for learning new concepts. Current deep learning methods are inefficient in remembering old concepts after being fed with new concepts, also widely know as catastrophic forgetting \cite{mccloskey1989catastrophic,kemker2017measuring}. Deep neural networks (DNNs) trained in an end-to-end fashion also has difficulty in learning new tasks in a sample efficient manner \cite{finn2017model}. It is conjectured that the cause of catastrophic forgetting and inefficiency in learning new tasks is from the stability-plasticity dilemma \cite{abraham2005memory}. Stability is required so that previously learned information can be retained through the limitation of abrupt weight changes. Plasticity on the other hand encourages large weight changes, resulting in the fast acquisition of new concepts with the trade-off of forgetting old concepts.

It is believed that by scaling up the currently available architecture, DNNs are able to generalize better \cite{brown2020language,radford2019language,devlin2018bert}. Tremendous effort is placed into neural architecture search (NAS) \cite{lee2020neuralscale,zoph2016neural,tan2019efficientnet,pham2018efficient,liu2018progressive} with the hypothesis that improvements on a structural level introduce inductive bias that improves the generalizability of a neural network. As most of the prior arts are evaluated on benchmark datasets that are distributed similarly to the training set that it is trained on, the evaluation results are not a good measure of the generalization. We argue that the ability to adapt, acquire new knowledge and recall previously learned information plays an important role in reaching true generalization. The importance of learning to learn, i.e.\ meta-learning, has shone the spotlight on two major research direction that we will focus on --- \textit{few-shot learning} and \textit{continual learning}. In few-shot learning \cite{finn2017model,nichol2018first,snell2017prototypical,gidaris2019boosting}, the goal is to learn novel concepts with as few samples as possible, i.e.\ evaluating the capability of adapting to new tasks. Whereas in continual learning, the ability to learn an increasing amount of concepts while not forgetting old ones is evaluated.

Following OML \cite{javed2019meta}, we separate the feature extraction part and the decision making part of the network, defined in OML as representation learning network (RLN) and prediction learning network (PLN) respectively. The fast and slow learning in OML is performed on an architecture level, i.e.\ RLN is updated in the outer loop (slow weights) and PLN is updated in the inner loop (fast weights). Such approach has proven to be helpful in learning sparse representation that are beneficial for fast adaptation and prevention of catastrophic forgetting. We take one step further by introducing sparsity on an architectural level, accomplished through the introduction of Attentive Independent Mechanisms (AIM). AIM is composed of a set of mechanisms that competitively attend to the input representation, having mechanisms that are closely related to the input representation being activated during inference. AIM can be understood as a mixture of experts competing to explain an incoming representation, hence only the mechanisms that best explain the input representation will be updated, leading to a sparse representation or modeling on an architectural level. Having sparse modeling on an architectural level for higher-order representations has its benefits, as only the experts or mechanisms that best explain a task will be involved in the learning process, helping in the acceleration of learning new concepts and the mitigation of catastrophic forgetting. To demonstrate the potential of AIM as a fundamental building block for fast learning without forgetting, we demonstrate its strength on few-shot classification \cite{finn2017model,rusu2018meta,zintgraf2018fast} and continual learning \cite{beaulieu2020learning,javed2019meta,kemker2017measuring} benchmarks. 

Our contributions are as follows:
    (1) In Section \ref{sec:method}, we give a detailed description and formulation of AIM --- a novel module that can be used for few-shot and continual learning.
    (2) We apply AIM on few-shot learning and continual learning tasks in Section \ref{sec:fewshot} and Section \ref{sec:continual} respectively. Qualitative and quantitative results are shown for both learning tasks, giving readers an insight on the importance of having AIM in the context of few-shot and continual learning. For few-shot classification, experiments are performed on CIFAR-FS and MiniImageNet whereas for continual learning, experiments are performed on Omniglot, CIFAR-100 and MiniImageNet. Substantial improvement in accuracy over prior arts are shown.


\section{Related Work}

Meta-learning revolves on the idea of learning to learn, hoping that through the observation of training iterations on a few tasks, we are able to generalize to unseen tasks with only a few or zero samples. Meta-learning is usually composed of a support set and a query set. The support set is used for fast adaptation and the query set is used to evaluate the adapted model and to meta-learn the adaptation procedure. Model-based meta-learning methods include the work by \cite{mishra2017simple} that uses a meta-learner based on a LSTM \cite{hochreiter1997long} which includes all previously seen samples, i.e.\ all support samples of a task are considered during the class prediction of query samples through an attentive mechanism. Another similar work by \cite{santoro2016meta} augments LSTM with an external memory bank. \cite{munkhdalai2017meta} incorporates fast and slow weights for few-shot classification. 

Metric-based meta-learning methods include Siamese Network proposed by \cite{koch2015siamese} which predicts whether two images originate from the same class. \cite{vinyals2016matching} proposed Matching Networks that uses cosine distance in an attention kernel to measure the similarity of images in its embedding space. \cite{snell2017prototypical} later found that using Euclidean distance as a metric instead of cosine distance improves performance. A generalization of all the mentioned work is done by modeling the metric using a graph neural network proposed by \cite{garcia2017few}.

Optimization-based meta-learning includes \cite{ravi2016optimization} that proposed using a LSTM meta-learner which provides gradient to a convolutional network-based fast learner. \cite{finn2017model,nichol2018first} proposed an inner and outer-loop optimization method having fast adaptation in the inner-loop and an outer-loop update that backpropagates through the inner-loop updates. \cite{zintgraf2018fast} used the concept of inner and outer loop-update by having the context parameters (embeddings of tasks) updated in the inner-loop. LEO \cite{rusu2018meta} has its classifier weights generated by a low-dimensional latent embedding updated in the inner-loop. \cite{gidaris2018dynamic} proposed a similar approach where classification weights are generated using feature vectors that corresponds to the support set. SIB \cite{Hu2020Empirical} performs transductive inference using synthetic gradient \cite{jaderberg2017decoupled} on the feature averaging variant classifier proposed by \cite{gidaris2018dynamic}. Transductive inference was first introduced to the context of few-shot classification by \cite{liu2018learning}, having a graph constructed for the support set and the query set, with labels propagated within the graph. As the architecture proposed by \cite{liu2018learning} is restrictive, \cite{hou2019cross} proposed a more general approach that uses a cross attention module that models semantic relevance between the support and query set.

In continual learning, the objective is to mitigate catastrophic forgetting \cite{kemker2017measuring}. Earlier works are based on regularization method, with \cite{hinton1987using} proposing the use of fast and slow training weights, borrowing the idea of plasticity and stability for network training. This idea is then adopted by OML \cite{javed2019meta} to learn representations that are useful for future learning and helps in mitigating catastrophic forgetting. Similarly, fast and slow learning is applied to ANML \cite{beaulieu2020learning}, having a neuromodulatory network modeled using slow weights. \cite{abati2020conditional} uses task-specific gate module and prediction head to reduce competitive effect between classes. 
A criterion is designed in \cite{aljundi2019online} to store most-interfered samples in a fixed-sized rehearsal memory.

\section{Method}\label{sec:method}
\begin{figure*}
    \centering
    \includegraphics[width=\textwidth]{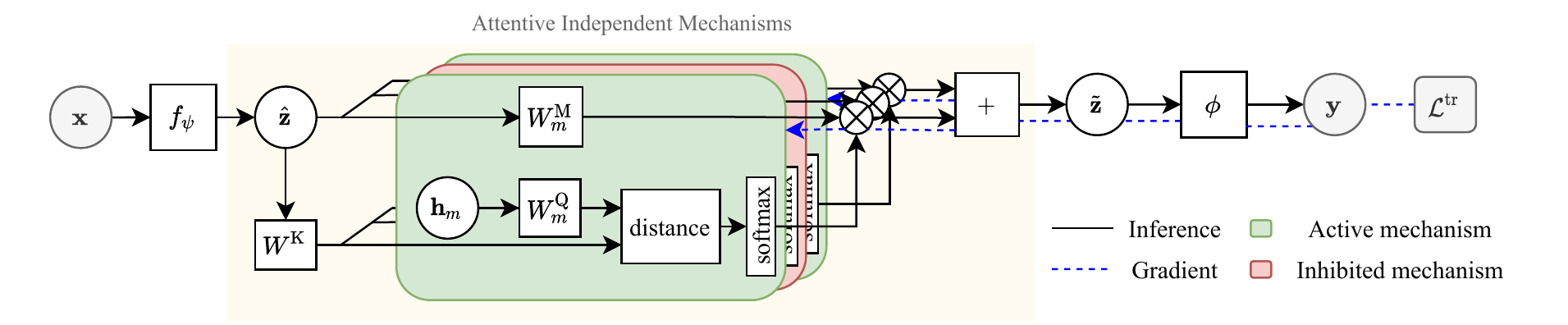}
    \caption{AIM is inserted right after the feature extractor $f_{\boldsymbol{\psi}}$ and before the output classifier $\boldsymbol{\phi}$. Only mechanisms closely related to the input representation are active (green boxes) and updated during the training phase (blue dashed lines).}
    \label{fig:AIM}
\end{figure*}
\begin{figure}[!tbp]
\centering
  	\subfloat[Synthetic Information Bottleneck (SIB) \cite{Hu2020Empirical}]{\includegraphics[width=0.48\textwidth]{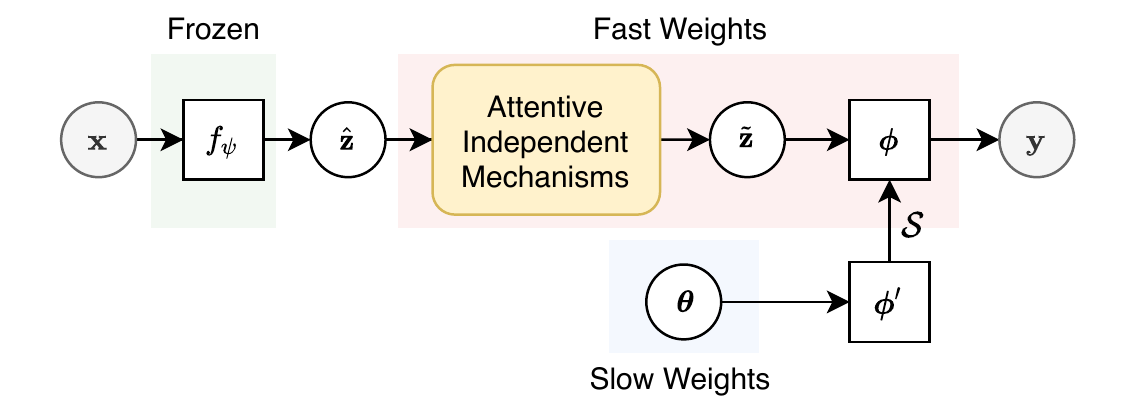}\label{fig:sib_aim}}
	\hfill
  	\subfloat[Online aware Meta-Learning (OML) \cite{javed2019meta}]{\includegraphics[width=0.48\textwidth]{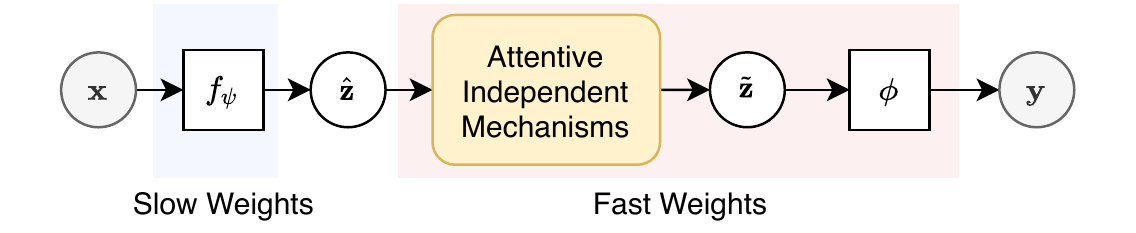}\label{fig:oml_aim}}
	\hfill
  	\subfloat[A Neuromodulated Meta-Learning Algorithm (ANML) \cite{beaulieu2020learning}]{\includegraphics[width=0.48\textwidth]{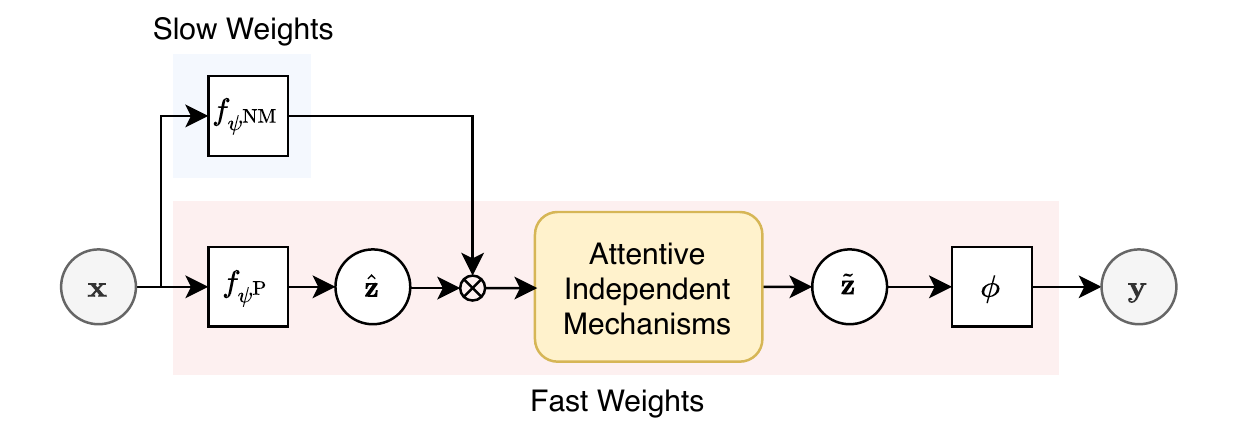}\label{fig:anml_aim}}
\caption{Applying AIM on both few-shot learning (\protect\subref{fig:sib_aim} SIB) and continual learning (\protect\subref{fig:oml_aim} OML and \protect\subref{fig:anml_aim} ANML) frameworks. For all frameworks, AIM (yellow) is placed directly after the feature extractor, $f_{\boldsymbol{\psi}}(\cdot)$. With different learning scheme (fast and slow) used in meta-learning, weights or modules that correspond to fast update are highlighted in red, slow update are in blue and frozen weights are in green. }
\label{fig:applying_aims}
\end{figure}

As Attentive Independent Mechanisms (AIM) is used to model higher order information, we place it right after a feature extractor, defined as $\mathbf{z}=f_{\boldsymbol{\psi}} (\mathbf{x})$. $f_{\boldsymbol{\psi}}  (\cdot)$ is a series of convolutional layers parameterized by $\boldsymbol{\psi}$, $\mathbf{x}$ is an input sample and $\mathbf{z}$ is its corresponding representation. AIM is a module that is parameterized by $\mathbf{W}$ and is defined as $\mathcal{A}_\mathbf{W}$. The representation from AIM is then fed to a linear layer $\boldsymbol{\phi}$ for the task of classification. An illustration of AIM as a module is shown in Figure \ref{fig:AIM}. We also show an illustration on the application of AIM to existing meta-learning frameworks used for few-shot learning and continual learning in Figure \ref{fig:applying_aims}. We first describe the implementation of AIM as module in Section \ref{sec:AIM} followed by its integration to SIB \cite{Hu2020Empirical} for few-shot learning in Section \ref{sec:sib_aim} and to OML \cite{javed2019meta} and ANML \cite{beaulieu2020learning} for continual learning in Section \ref{sec:fast_and_slow_aim}.

\subsection{Attentive Independent Mechanisms}\label{sec:AIM}

The goal of AIM is to learn a sparse set of mechanisms, i.e.\ mixture of experts, to decouple the modeling of higher order information from the feature extraction pipeline. These mechanisms compete and attend to the input representation in a top-down fashion using cross-attention \cite{lee2018stacked,lee2019set}. Through the strict selection of mechanisms, a sparse set of mechanisms will be selected for every task, inducing an architectural bias that helps in fast adaptation to new tasks and mitigating catastrophic forgetting. 
The structure of AIM is composed of a set of independent mechanisms, each parameterized by its own set of parameters. Each mechanism acts as an independent expert that collaborate with other experts to solve a particular task. AIM can be viewed as a static version of RIMs \cite{goyal2019recurrent}, i.e.\ temporal modeling of hidden states using LSTM \cite{hochreiter1997long} found in RIMs is removed.
For RIMs, the model is fed with a continuous stream of inputs, making dynamical modeling using LSTM intuitive. For AIM, the assumption of having continuous stream of inputs does not hold as the practice of few-shot classification and continual learning have i.i.d.\ data being fed into the model during training and inference. Departing from RIMs, the objective of AIM is to show that through a mixture of experts, new concepts can be easily learned with minimal catastrophic forgetting.
We hypothesize that by having a set of independent mechanisms, a sparse set of factorized representations or concepts can be extracted from the input representation. Such concepts have properties that are tasks-invariant which can be helpful in learning new tasks. 
The learning of concepts in AIM can also be understood as the amortized version of memory based models that stores samples either in the form of images or representations \cite{santoro2016meta}, which scales with the size of tasks in the system without limitation. AIM on the other hand performs implicit modeling of samples, analogous to the amortized modeling using a DNN instead of using a non-parametric method that stores samples from the training set for inference \cite{chang2011libsvm}.

Following RIMs, AIM has a null vector $\mathbf{\emptyset}$ that is concatenated with the input representation $\mathbf{z}$, giving us $\hat{\mathbf{z}}=[\mathbf{z}^T, \mathbf{\emptyset}^T]^T$. The mechanisms then attend to the incoming latent representation $\hat{\mathbf{z}}$ as:
\begin{equation}
    \tilde{\mathbf{z}} = \hat{\mathbf{z}}\left(\sum_{m=1}^{M}w_{m}(\hat{\mathbf{z}})W_{m}^{\text{M}}\right), \label{eq:weighted}
\end{equation}
which could be understood as the passing of input representation $\hat{\mathbf{z}}$ through the weighted-summation of the mechanism weights, $W_{m}^{\text{M}}$. The summation of the outputs of the mechanisms makes the extension to arbitrary number of mechanisms trivial when compared to the concatenation of outputs used in RIMs. Concatenation is also infeasible when the output dimension of $W_m^{\mathrm{M}}$ is large, resulting in a wide input dimension for the upcoming layer. The summation of mechanisms also has the property of permutation invariance, reducing the complexity of the output classifier $\boldsymbol{\phi}$.

To encourage sparsity, we enforce the mechanisms to compete with each other to attend to the incoming representation. This is done by having only the weights of mechanisms that are closely related to the input representation to be selected, i.e.\ only top $K$ mechanisms out of a total of $M$ mechanisms are selected for the downstream prediction tasks. The strict selection of mechanisms forces the mechanisms to compete with each other to attend to the incoming signal, modeling the \textit{biased competition theory} of selective attention \cite{desimone1995neural}. The selection of mechanisms is given as:
\begin{equation}
w_{m}(\hat{\mathbf{z}}) = \begin{cases} \widetilde{w}_{m}(\hat{\mathbf{z}}), & \text{if } m \in \mathrm{top}_{K}\left(\widetilde{w}_{1}(\hat{\mathbf{z}}),\hdots,\widetilde{w}_{M}(\hat{\mathbf{z}})\right), \\
0, & \text{otherwise.} \end{cases}    \label{eq:topk}
\end{equation}
The indices corresponding to the top $K$ values from a set is returned by the $\mathrm{top}_{K}(\cdot)$ operator. The weights used to weight the importance of the selected mechanisms are composed of the softmax of the normalized inner-product, $\langle \cdot, \cdot \rangle$, between the mechanisms' hidden state $\mathbf{h}_m$ and the input representation $\mathbf{z}$ that are first mapped to a lower-dimensional embedding by the query weight $W_m^{\text{Q}}$ and key weight $W^{\text{K}}$ of output dimension $d$ respectively, given as:
\begin{equation}
\widetilde{w}_{m}(\hat{\mathbf{z}}) = \mathrm{softmax}\left(\frac{\langle \mathbf{h}_{m}W_{m}^{\mathrm{Q}}, \hat{\mathbf{z}} W^{\mathrm{K}} \rangle}{\sqrt{d}}\right).\label{eq:softmax}
\end{equation}
Note that $\mathrm{softmax}$ is applied locally for each mechanism, i.e.\ the transformation of the attention values corresponding to $\mathbf{z}$ and $\emptyset$ into a probabilistic one. The value that corresponds to the input (not null) dimension from (\ref{eq:softmax}) is used for the top $K$ comparison in (\ref{eq:topk}).




\paragraph{Intervention during training.}
The training of AIM can be understood as an intervention procedure with the model selecting a few mechanisms to be included during the forward pass phase of training. 
Mechanisms that perform well on the training data are rewarded by having gradient update directed to the activated mechanisms, with the sensitivity to novel inputs reflected on $\mathbf{h}_m$. As one can predict, there is a possibility of the occurrence of \textit{mechanism-overfitting}, where only a fixed set of mechanisms are active for all training tasks, losing the original motivation of having a sparse set of mechanisms acting as experts on different tasks. Mechanism-overfitting is also equivalent to having a DNN with multiple residual paths, resembling a single layer of Inception \cite{szegedy2015going}, diverting from our original goal of building models that are invariant across tasks.

To prevent the collapsing towards having only a few active mechanisms for all tasks, the trick is to enable the exploration of different amount of mechanisms during training, instead of locking down to the top $K$ mechanisms. Stochasticity is introduced into the selection process by sampling top $K+l$ (also known as \textit{stochastic sampling count}) instead of top $K$ mechanisms. We then perform uniform sampling without replacement of $K$ mechanisms from the top $K+l$ mechanisms, where the original sampling condition of (\ref{eq:topk}) can now be written as:

\begin{equation}
w_{m}(\hat{\mathbf{z}}) = \begin{cases} \widetilde{w}_{m}(\hat{\mathbf{z}}), & \text{if } m \in \{\mathcal{K} \subseteq S \mid \vert\mathcal{K}\vert = K \} , \\
& \text{s.t. } S = \mathrm{top}_{K+l}\left(\widetilde{w}_{1}(\hat{\mathbf{z}}),\hdots,\widetilde{w}_{M}(\hat{\mathbf{z}})\right) \\
0, & \text{otherwise.} \end{cases}    \label{eq:topk_stochastic}
\end{equation}
Here, $|\cdot|$ is the cardinality operator to ensure that the sampled subset $\mathcal{K}$ is of size $K$ and is sampled without replacement. Such intervention is analogous to stochastic intervention \cite{korb2004varieties} and dropout \cite{srivastava2014dropout} which adds stochasticity to the training of AIM, preventing the locking down to a few mechanisms that are attended to upon initialization.

\paragraph{Training and evaluation of AIM.} 
Weight updates in AIM is similar to a typical layer in DNNs, i.e.\ gradients are backpropagated from the final loss function. A distinct difference from a conventional module in DNNs is that only the mechanisms activated during a forward pass are updated, resulting in a sparse set of weight updates. As AIM is designed to model higher order concepts, it is placed in the higher level of a DNN and has fast weights that are updated in the inner-loop of a meta-learning pipeline. The role of AIM as a module is shown in Figure \ref{fig:AIM}. The procedure for the meta-training of AIM for both few-shot learning and continual learning is shown in Algorithm \ref{alg:training}, whereas the meta-testing counterpart is shown in Algorithm \ref{alg:testing}. The algorithms shown are applicable for both few-shot and continual learning with the distinction between both highlighted with different colors --- few-shot learning using SIB in {\color{ForestGreen}green} and continual learning using OML and ANML in {\color{blue}blue}. Step sizes for the inner-loop and outer-loop are defined as $\nu^{\text{in}}$ and $\nu^{\text{out}}$ respectively. Step size for synthetic gradient update used for SIB is defined as $\epsilon$. For few-shot learning, the fast adaptation of AIM is evaluated using the meta-testing test set of the sampled task, i.e.\ $\mathcal{S}_\text{test}$ in the outer-loop. For continual learning, evaluation is performed after the completion of meta-training, and is tested on the entire meta-test train set $\mathcal{S}'_\text{train}$ and meta-test test set $\mathcal{S}'_\text{test}$.


\begin{algorithm}[tb]
   \caption{Meta-Training: Training of AIM}
   \label{alg:training}
\begin{algorithmic}[1]
   \Require $N$ sequential tasks $\mathcal{T}$; step size $\nu^{\mathrm{in}},\nu^{\mathrm{out}},\epsilon$; inner iterations $T$; modules $f_\psi, \mathcal{A}_{\mathbf{W}}, \boldsymbol{\phi}, \boldsymbol{\theta}$ (SIB only)
\While{not done}
    \State {\small $\{\mathcal{S}_{\text{train}},\mathcal{S}_{\text{test}}\} \sim \mathcal{T} $ \Comment{ {\color{ForestGreen}SIB: i.i.d.}; {\color{blue}continual: sequential} }}
   \For{$t\gets 1, T$} 
      \State Update fast weights using $\mathcal{S}_{\text{train}}$\Comment{step size: $\nu^{\mathrm{in}}$}
      \Statex \qquad {\color{ForestGreen}SIB: $\mathcal{A}_\mathbf{W}$} \quad {\color{blue}OML: $\mathcal{A}_\mathbf{W},\boldsymbol{\phi}$\quad ANML: $f_{\boldsymbol{\psi}^{\text{P}}},\mathcal{A}_\mathbf{W}, \boldsymbol{\phi}$}
   \EndFor
   \State {\color{ForestGreen}Update $\boldsymbol{\phi}$ using transductive inference \Comment{step size:$\epsilon$}}
    \State Update slow weights using $\mathcal{S}_{\text{test}}$\Comment{step size: $\nu^{\mathrm{out}}$}
      \Statex $\qquad\quad$ {\color{ForestGreen}SIB: $\boldsymbol{\theta}$}$\quad$ {\color{blue}OML: $f_{\boldsymbol{\psi}}\quad$ ANML: $f_{\boldsymbol{\psi}^{\mathrm{NM}}}$}
\EndWhile
\end{algorithmic}
\end{algorithm}

\begin{algorithm}[tb]
   \caption{Meta-Testing: Evaluation of AIM}
   \label{alg:testing}
\begin{algorithmic}[1]
   \Require $N$ sequential unseen tasks $\mathcal{T}$; step size $\nu^{\mathrm{in}}, \epsilon$; inner iterations $T$; modules $f_\psi, \mathcal{A}_{\mathbf{W}}, \boldsymbol{\phi}, \boldsymbol{\theta}$ (SIB only)
   
  \Statex {\color{blue}$\mathcal{S}'_\text{train}=\{\}$; $\mathcal{S}'_\text{test}=\{\}$ \Comment{initialize empty set}}
\For{$n\gets 1, N$} 
    \State {\small $\{\mathcal{S}_{\text{train}},\mathcal{S}_{\text{test}}\} \sim \mathcal{T}_n $ \Comment{ {\color{ForestGreen}SIB: i.i.d.}; {\color{blue}continual: sequential} }}
    \State{\small\color{blue} $\mathcal{S}'_\text{train},\mathcal{S}'_\text{test}=\{\mathcal{S}'_\text{train}, \mathcal{S}_\text{train}\},\{\mathcal{S}'_\text{test}, \mathcal{S}_\text{test}\}$ \Comment{store trajectory}}
   \For{$t\gets 1, T$} 
      \State Update fast weights using $\mathcal{S}_{\text{train}}$\Comment{step size: $\nu^{\mathrm{in}}$}
      \Statex \qquad {\color{ForestGreen}SIB: $\mathcal{A}_\mathbf{W}$} \quad {\color{blue}OML: $\mathcal{A}_\mathbf{W},\boldsymbol{\phi}$\quad ANML: $f_{\boldsymbol{\psi}^{\text{P}}},\mathcal{A}_\mathbf{W}, \boldsymbol{\phi}$}
   \EndFor
   \State {\color{ForestGreen}Update $\boldsymbol{\phi}$ using transductive inference\Comment{step size:$\epsilon$}}
   \State {\color{ForestGreen}Evaluate on $\mathcal{S}_{\text{test}}$}
\EndFor
\State {\color{blue}Evaluate on $\mathcal{S}'_{\text{train}}$ \Comment{end of meta-test training trajectory}}
\State {\color{blue}Evaluate on $\mathcal{S}'_{\text{test}}$ \Comment{eval on entire meta-test testing set}}
\end{algorithmic}
\end{algorithm}

\subsection{Few-Shot Learning Using SIB}\label{sec:sib_aim}
SIB is composed of two works: synthetic gradient modeling \cite{jaderberg2017decoupled} and a feature averaging classifier \cite{gidaris2018dynamic}. In \cite{jaderberg2017decoupled}, the idea is to use a synthetic gradient model, $\mathcal{S}$, that is meta-learned to generate gradient when labeled data is absent for transductive inference, i.e.\ update of weights without gradients propagated from a loss that is dependent on label. In \cite{gidaris2018dynamic}, a classifier is defined as the cosine similarity between feature representations $\tilde{\mathbf{z}}$ and classification weight vectors $\boldsymbol{\phi}$. $\boldsymbol{\phi}$ is generated using an external classification weight generator $G_{\boldsymbol{\theta}}(\cdot)$ parameterized by $\boldsymbol{\theta}$ followed by iterative update by the synthetic gradient model $\mathcal{S}$. Feature vectors of $P$ training samples of a novel category $\bar{\mathbf{Z}}= \{\bar{\mathbf{z}}^{(i)}\}_{i=1}^P$ are fed as input to generate a new set of weights for classification, $\boldsymbol{\phi}'=G_{\boldsymbol{\theta}}(\bar{\mathbf{Z}})$. In SIB, feature averaging based weight inference is used, i.e.\ the classification weight vector is obtained as $\boldsymbol{\phi}' = \boldsymbol{\theta} \odot \mathbf{w}_\mathrm{avg}$, where $\odot$ is the Hadamard product and $\mathbf{w}_\mathrm{avg} = \frac{1}{P} \sum_{i=1}^P \bar{\mathbf{z}}^{(i)}$ ($\bar{\mathbf{z}}$ is the $\ell_2$-normalized version of $\tilde{\mathbf{z}}$). The classification weight vector $\boldsymbol{\phi}'$ is then updated iteratively using the synthetic gradient model in SIB, given as $\boldsymbol{\phi} = \mathcal{S}(\boldsymbol{\phi}')$. Both the synthetic gradient model and the weights of the weight generator $\boldsymbol{\theta}$ are meta-learned, i.e.\ updated in the outer-loop. To encourage sparse modeling of higher order concepts in the network, AIM is inserted right after the feature extractor $f_{\boldsymbol{\psi}}(\cdot)$ and before the output linear classifier $\boldsymbol{\phi}$ that is generated using $G_{\boldsymbol{\theta}}(\bar{\mathbf{Z}})$ and $\mathcal{S}$, or,
\begin{equation}
\mathbf{y} = \boldsymbol{\phi}\left( \mathcal{A}_{\mathbf{W}} \left( f_{\boldsymbol{\psi}} (\mathbf{x}) \right) \right),\quad \text{where } \boldsymbol{\phi}= \mathcal{S}\left( G_{\boldsymbol{\theta}}\left(f_{\boldsymbol{\psi}}(\bar{\mathbf{X}})\right)\right).
\end{equation}

\paragraph{Training.} Following the training pipeline in SIB \cite{Hu2020Empirical}, the weights of the feature extractor $\boldsymbol{\psi}$ are frozen to simplify the training procedure. The weights of the AIM, $\mathbf{W}$, and the output linear classifier, $\boldsymbol{\phi}$, are updated as fast weights, i.e.\ inner-loop. Only the weights of the classification weight generator $\boldsymbol{\theta}$ are updated as slow weight, i.e.\ outer-loop. The application of AIM to SIB is shown in Figure \ref{fig:sib_aim}.

\subsection{Continual Learning: Learning Fast and Slow}\label{sec:fast_and_slow_aim}
It is shown in the task of continual learning that learning fast and slow from the context of meta-learning is helpful for the mitigation of catastrophic forgetting \cite{javed2019meta,beaulieu2020learning}. OML \cite{javed2019meta} and ANML \cite{beaulieu2020learning} are example frameworks for continual learning that uses this methodology, showing promising results. To validate our claim on the importance of incorporating sparse modeling on an architectural level for the mitigation of catastrophic forgetting, we insert AIM into both OML and ANML and observe the resulting performance.

\paragraph{OML.} The entire architecture is split into two parts --- \textit{representation learning network} (RLN) and \textit{prediction learning network} (PLN). RLN uses slow weights and PLN uses fast weights. Following our notations, RLN is the feature extractor in our work, $f_{\boldsymbol{\psi}}$, and PLN is the classifier (not limited to a single layer), $\boldsymbol{\phi}$, in our work. AIM is inserted after the RLN and before the PLN, or,
\begin{equation}
\mathbf{y} = \boldsymbol{\phi}\left(\mathcal{A}_{\mathbf{W}}(f_{\boldsymbol{\psi}}(\mathbf{x}))\right).
\end{equation}
AIM is trained jointly with PLN, i.e.\ they have fast weights. The application of AIM to OML is shown in Figure \ref{fig:oml_aim}.

\paragraph{ANML.} Two set of feature extractors are used in ANML --- a \textit{neuromodulatory network}, $f_{\boldsymbol{\psi}^{\mathrm{NM}}}$, and a \textit{prediction network}, $\boldsymbol{\phi}\cdot f_{\boldsymbol{\psi}^{\mathrm{P}}}$. The role of the neuromodulatory network is to modulate the latent representation of the prediction network, i.e.\ the output of $f_{\boldsymbol{\psi}^\mathrm{P}}$ in Figure \ref{fig:anml_aim}. The output of the neuromodulatory network is element-wise multiplied with the outout of $f_{\boldsymbol{\psi}^\mathrm{P}}$ before passing to the final classifier, or $\mathbf{y} = \boldsymbol{\phi}\left(f_{\boldsymbol{\psi}^{\mathrm{NM}}}(\mathbf{x}) \odot f_{\boldsymbol{\psi}^{\mathrm{P}}}(\mathbf{x})\right)$. Only the neuromodulatory network has slow weights and the entire prediction network has fast weights. Similar to SIB and OML, AIM is inserted right after the feature extractor and uses fast weights,
\begin{equation}
\mathbf{y} = \boldsymbol{\phi} \left(\mathcal{A}_{\mathbf{W}} \left(f_{\boldsymbol{\psi}^{\mathrm{NM}}}(\mathbf{x}) \odot f_{\boldsymbol{\psi}^{\mathrm{P}}}(\mathbf{x})\right) \right).
\end{equation}


\section{Experiments}\label{sec:exp}
\subsection{Few-Shot Learning}\label{sec:fewshot}
\paragraph{Datasets.} 
For all datasets, class splits are disjoint. MiniImageNet \cite{vinyals2016matching} contains a total of 100 classes which are split into 64 training, 16 validation and 20 testing classes; images are of size $84\times84$. CIFAR-FS \cite{bertinetto2018meta} is created by dividing CIFAR-100 into 64 training, 16 validation and 20 testing classes; images are of size $32\times32$. For few-shot classification, each task (episode) consists of a train set and a test set. For each task, $k$ classes are sampled from the class pool mentioned. For each class, $n$ examples are drawn and are relabeled as $k$ disjoint classes forming the train set. For the test set, 15$k$ samples are used. We show results of $k=5$ for both $n=1$ and $n=5$.

\paragraph{Network architecture.}
We follow the setup in \cite{Hu2020Empirical,gidaris2018dynamic,qiao2018few,gidaris2019boosting} by using a 4-layer convolutional network with 64 feature channels (Conv-4-64) or a WideResNet (WRN-28-10) \cite{zagoruyko2016wide} as our feature extractor, $f_\psi$. $f_\psi$ is pretrained in a typical end-to-end supervised learning fashion, i.e.\ the entire training set is used for batch update. Our classifier is adopted directly from \cite{Hu2020Empirical,gidaris2018dynamic} having $\boldsymbol{\phi} = G_{\boldsymbol{\theta}}'(\bar{\mathbf{Z}})$. For transductive inference \cite{Hu2020Empirical}, the synthetic gradient network is modeled by a MLP of 3 layers and hidden size 8$k$. Classification is done by using the cosine-similarity based classifier found in \cite{Hu2020Empirical,gidaris2018dynamic}. For AIM, all weights are linear layers. The hidden state $\mathbf{h}_m$ of the mechanisms are of dimension 256. The key and query weights ($W^{\mathrm{K}}$ and $W^{\mathrm{Q}}_m$) maps the input and hidden state to a dimension of 128 to perform distance measurement. For the output dimension of the mechanism weights, $W^{\mathrm{M}}_m$, we picked 400 for CIFAR-FS trained on Conv-4-64 and 800 for the rest; this decision is based on the dimension of the flattened feature map at the output of the feature extractor (not cherry-picked).

\paragraph{Training details.} We use $M=32$ mechanisms with top $K=8$ mechanisms selected during inference with induced stochasticity by having $l=2$ during training. SGD is used a batch size of 1 for 50,000 steps with learning rate $\epsilon = 10^{-3}$ for SIB's classifier synthetic update, $\nu^{\mathrm{out}} = 5\times10^{-3}$ for outer-loop update and $\nu^{\mathrm{in}} = 3\times10^{-3}$ for inner-loop update. The feature extractor is frozen during training. 1,000 tasks are sampled from the validation set for hyperparameter selection at each training epoch. All experiments are run on a single GTX1080Ti using PyTorch. A complete run of Conv-4-64 on CIFAR-FS and WRN-28-10 on MiniImageNet takes less than 2 hours and 5 hours respectively.

\begin{figure}[!tbp]
\centering
  	\subfloat[SIB + AIM using Conv-4-64 on CIFAR-FS, 1-shot.]{\includegraphics[width=0.5\textwidth]{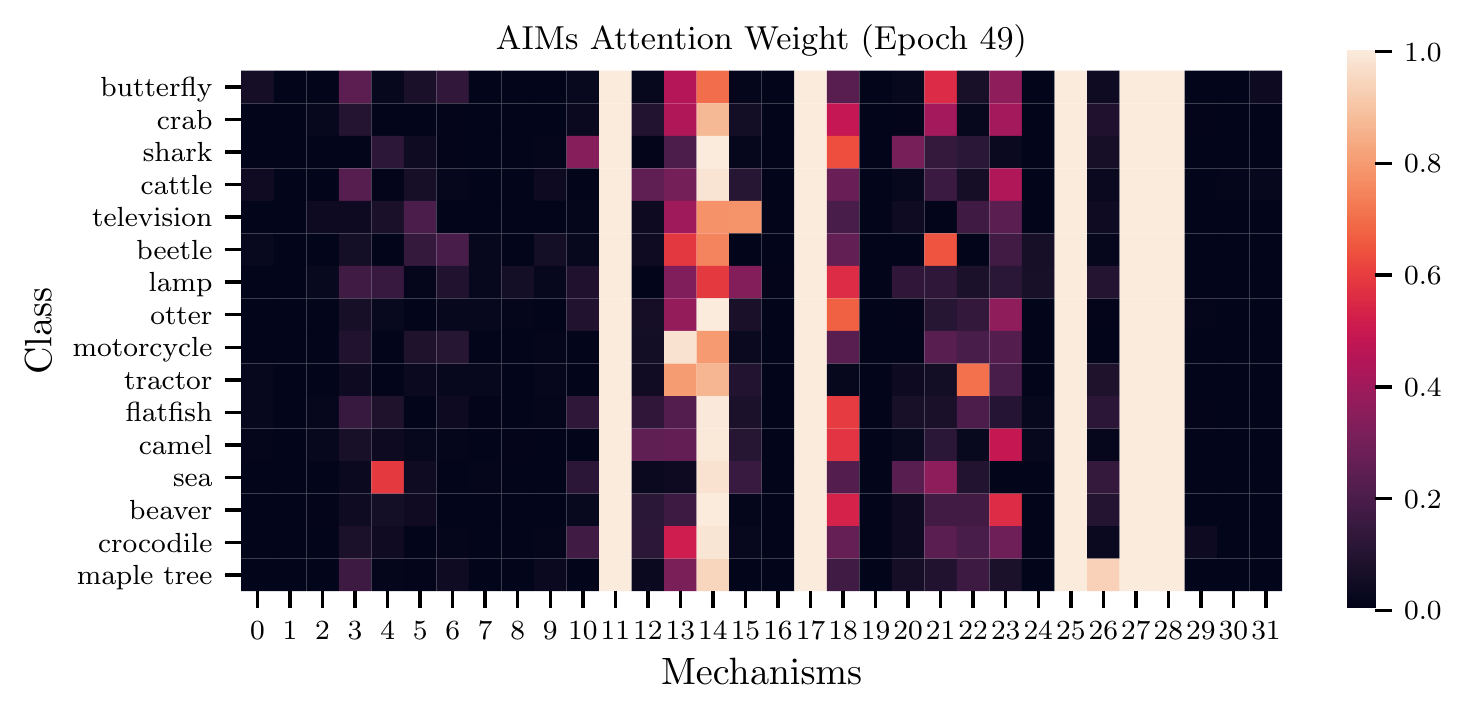}\label{fig:latency_vgg11_c10}}
	\hfill
  	\subfloat[OML + AIM on CIFAR-100.]{\includegraphics[width=0.5\textwidth]{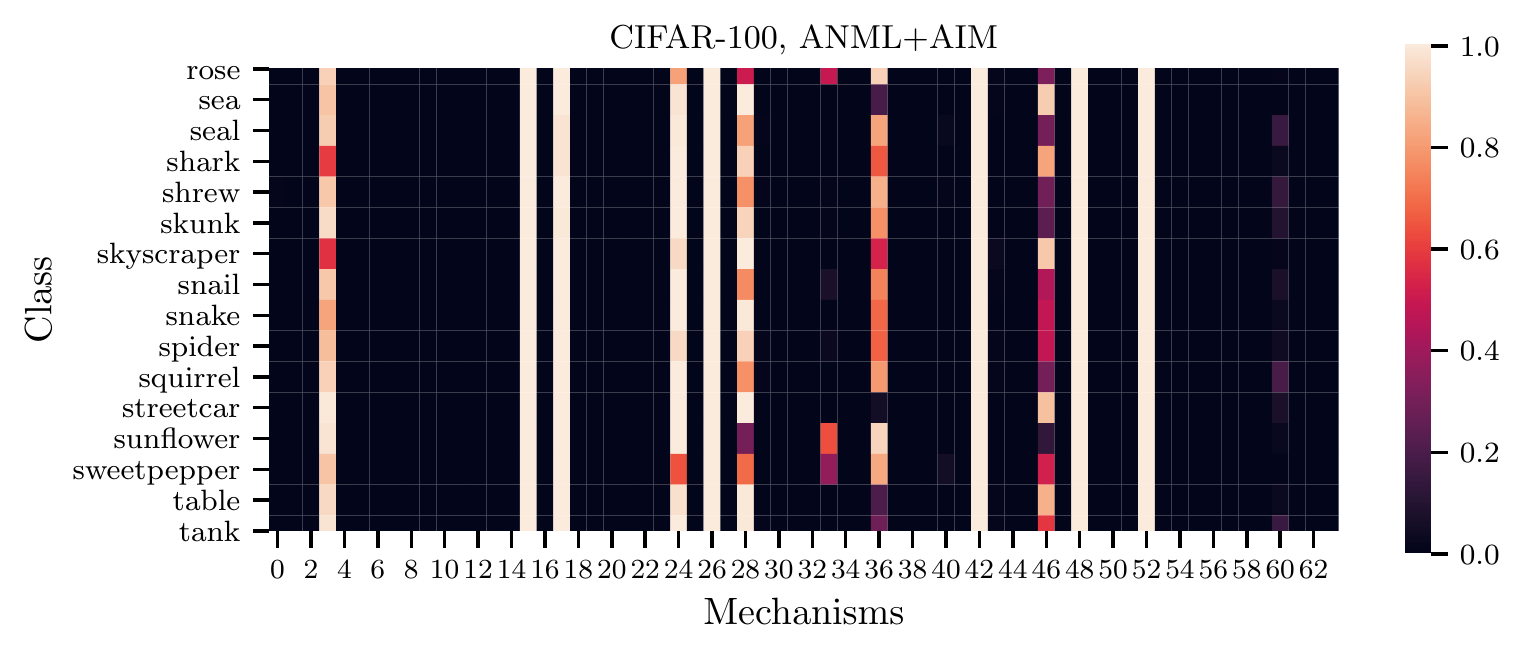}\label{fig:latency_vgg11_c10}}
\caption{With 1 indicating an active mechanism and 0 indicating an inhibited mechanism and having top $K$ mechanisms selected for every inference, the average of the activation for the same class across the entire validation set is taken here. The active mechanisms can be categorized into two sets: 1.\ fixed set of shared active mechanisms; 2.\ sparse set of mechanisms with class-dependent activations.
}
\label{fig:activation}
\end{figure}

\begin{table*}
\footnotesize
  \caption{Average classification accuracies with 95\% confidence intervals on the test-set of MiniImageNet and CIFAR-FS. 2000 episodes are sampled for MiniImageNet and CIFAR-FS using Conv-4-64 and WRN-28-10 as the feature extractor.}
  \label{tab:few-shot}
  \centering
  \begin{tabular}{lcccccc}
    \toprule
    \multirow{2}{*}{Method} & \multirow{2}{*}{Backbone} & \multirow{2}{*}{Transductive} & \multicolumn{2}{c}{MiniImageNet, 5-Way} & \multicolumn{2}{c}{CIFAR-FS, 5-Way}                   \\
     &  &  & 1-shot     & 5-shot & 1-shot     & 5-shot \\
    \midrule
    Matching Net \cite{vinyals2016matching}   & Conv-4-64 &     & 44.2\%  & 57\% & - & -    \\
    MAML \cite{finn2017model}  & Conv-4-64   &  & $48.7\pm1.8\%$  & $63.1\pm0.9\%$ & $58.9\pm1.9\%$ & $71.5\pm1.0\%$    \\
    Prototypical Net \cite{snell2017prototypical}  & Conv-4-64 &     & $49.4\pm0.8\%$  & $68.2\pm0.7\%$ & $55.5\pm0.7\%$ & $72.0\pm0.6\%$    \\
    Relation Net \cite{sung2018learning}  & Conv-4-64 &           & $50.4\pm0.8\%$ & $65.3\pm0.7\%$ & $55.0\pm1.0\%$ & $69.3\pm0.8\%$     \\
    TPN \cite{liu2018learning}  & Conv-4-64  &  \checkmark   & 55.5\% & 69.9\% & - & -      \\
    Gidaris \etal\cite{gidaris2019boosting} & Conv-4-64 &   & $54.8\pm0.4\%$ & $71.9\pm0.3\%$ & $63.5\pm0.3\%$ & $79.8\pm0.2\%$      \\
    SIB \cite{Hu2020Empirical}  &  Conv-4-64   & \checkmark & $58.0\pm0.6\%$ & $70.7\pm0.4\%$ & $68.7\pm0.6\%$ & $77.7\pm0.4\%$      \\
    SIB + Linear layer & Conv-4-64    & \checkmark  & $60.07\pm0.59\%$ & $73.70\pm0.38\%$ & $68.75\pm0.62\%$ & $79.99\pm0.39\%$      \\
    AIM (ours)  & Conv-4-64  & \checkmark   & $\mathbf{61.90\pm0.57\%}$ & $\mathbf{74.55\pm0.38\%}$ & $\mathbf{71.09\pm0.62\%}$ & $\mathbf{80.48\pm0.40\%}$      \\
    \midrule
    TADAM \cite{oreshkin2018tadam} & ResNet-12 & & $58.5\pm0.3\%$  & $76.7\pm0.3\%$ & - & -    \\
    SNAIL \cite{santoro2016meta}    & ResNet-12 &  & $55.7\pm1.0\%$  & $68.9\pm0.9\%$ & - & -    \\
    CTM \cite{li2019finding}   & ResNet-18 & \checkmark    & $64.1\pm0.8\%$ & $80.5\pm0.1\%$ & - & -      \\
    LEO \cite{rusu2018meta}     & WRN-28-10 &   & $61.8\pm0.1\%$ & $77.6\pm0.1\%$ & - & -      \\
    Gidaris \etal\cite{gidaris2019boosting} & WRN-28-10  & & $62.9\pm0.5\%$ & $79.9\pm0.3\%$ & $73.6\pm0.3\%$ & $86.1\pm0.2\%$      \\
    SIB \cite{Hu2020Empirical} & WRN-28-10   &  \checkmark &  $70.0\pm0.6\%$ & $79.2\pm0.4\%$ & $80.0\pm0.6\%$ & $85.3\pm0.4\%$      \\
    SIB + Linear layer &WRN-28-10   &  \checkmark &  $67.38\pm0.54\%$ & $80.54\pm0.34\%$ & $78.02\pm0.55\%$ & $79.91\pm0.38\%$      \\
    AIM (ours) & WRN-28-10   & \checkmark  &  $\mathbf{71.22\pm0.57\%}$ & $\mathbf{82.25\pm0.34\%}$ & $\mathbf{80.20\pm0.55\%}$ & $\mathbf{87.34\pm0.36\%}$      \\
    \bottomrule
  \end{tabular}
\end{table*}
\subsubsection{Qualitative Study: Activation of AIM}
We show heatmaps that illustrate mechanisms activated for different classes from the validation set in Figure \ref{fig:activation}. The heatmap is plotted by averaging the mechanisms' activity for each class over the entire validation set, with 1 and 0 indicating active and inhibited mechanism respectively. We can observe that there's a set of mechanisms that are shared among tasks and another set that are distributed sparsely. The sharing of mechanisms can be understood as different classes sharing similar concepts. The sparse allocation of mechanisms over different classes show that there are features that are invariant for certain classes only, improving resiliency to covariate shift among distributions. 

\begin{figure}[!tbp]

\centering
  	\subfloat{\includegraphics[width=0.235\textwidth]{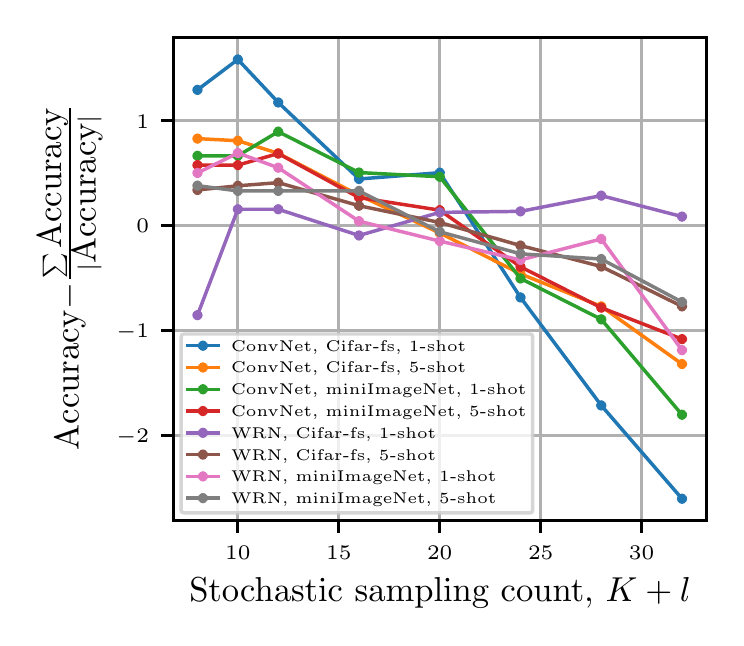}\label{fig:stochastic_sam}}
	\hfill
	\subfloat{\includegraphics[width=0.235\textwidth]{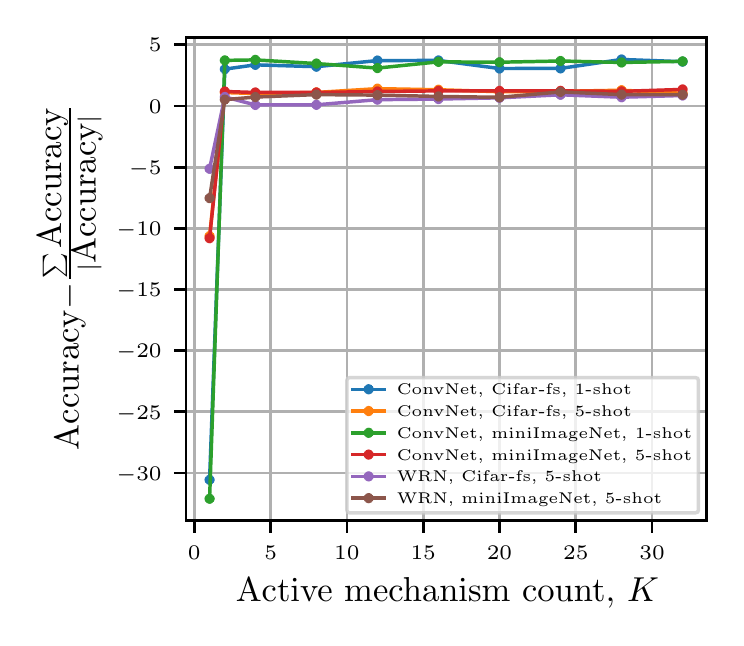}\label{fig:active_mech}}
\caption{Illustrates the accuracy obtained by varying \protect\subref{fig:stochastic_sam} stochastic sampling count ($K=8$ and $l$ is manipulated) and \protect\subref{fig:active_mech} active mechanisms count ($l=0$ and $K$ is manipulated). The zero mean-ed accuracy is shown to better demonstrate the change in accuracy across different model-dataset pairs. $|\cdot|$ is the cardinality operator.}
\label{fig:latency_comparison}
\end{figure}

\begin{figure*}[!tbp]

\centering
  	\subfloat{\includegraphics[width=0.3\textwidth]{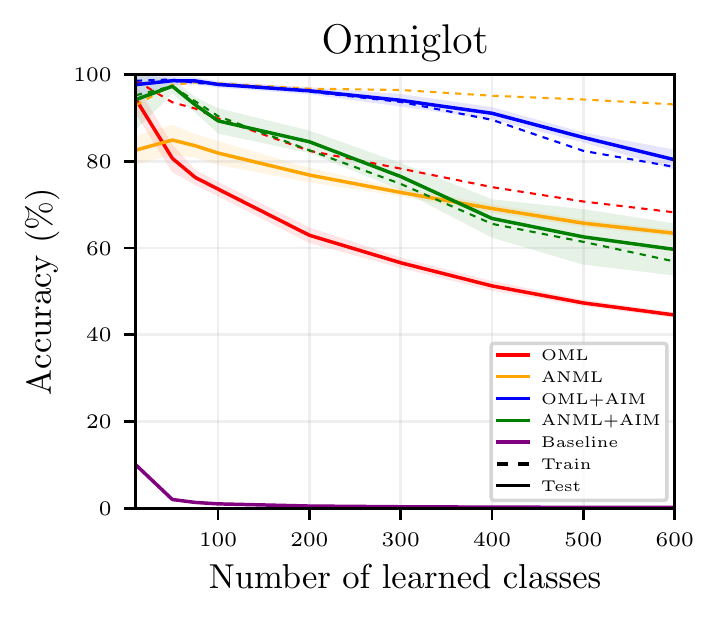}\label{fig:omniglot}}
	\hfill
	\subfloat{\includegraphics[width=0.3\textwidth]{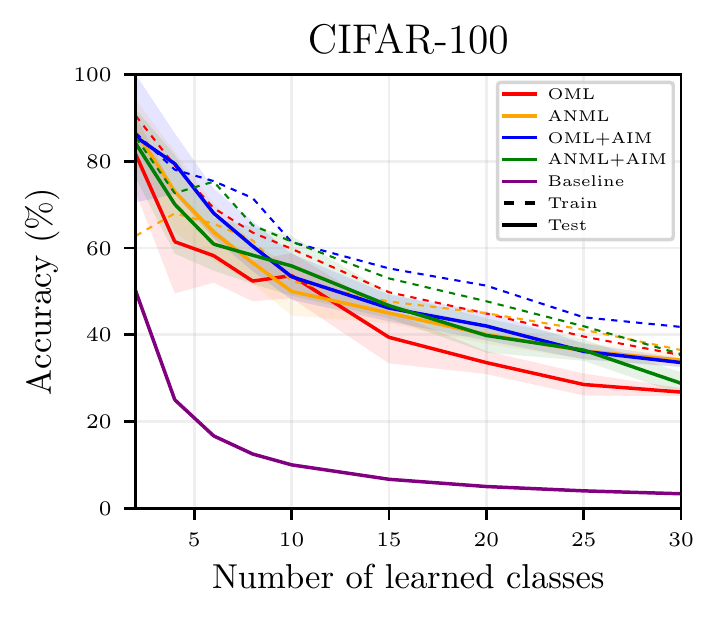}\label{fig:cifar100}}
	\hfill
	\subfloat{\includegraphics[width=0.3\textwidth]{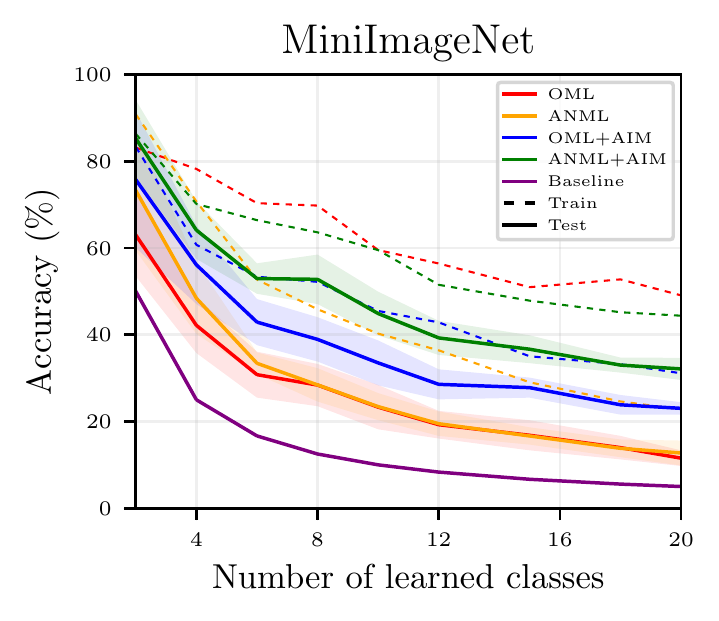}\label{fig:miniimagenet}}
\caption{Evaluation of continual learning methods using dataset of various scales. Meta-test testing (training) trajectories are shown in solid (dashed) lines. All curves are averaged over 10 runs with standard deviation shown.}
\label{fig:benchmark_continual}
\end{figure*}

\subsubsection{Quantitative Study}

\paragraph{Stochastic sampling count.} To show the importance of inducing stochasticity in the mechanism selection process for inference, we perform an empirical study by varying the stochastic sampling count, $K+l$. We fix $K=8$ and vary $l$ from 0 to 24. As we can see from Figure \ref{fig:stochastic_sam}, the accuracy obtained by varying $l$ have different maximum for different datasets, models and number of shots. For most cases, the peak accuracy usually occurs at small value of $l$ and slowly deteriorates as more stochasticity is introduced.

\paragraph{Number of active mechanisms.} An interesting question would be how maybe active mechanisms are required to reap the benefits of sparse activations. Empirical study is performed as shown in Figure \ref{fig:active_mech}, showing the accuracy obtained by varying the number of active mechanisms $K$ from 1 to 32. The results show that accuracy is low when $K$ is small and saturates for larger values of $K$. This shows that a limited set of active mechanisms is sufficient. Sparsity in representation can still be met when the number of active mechanisms is large, but it will be cost inefficient during both training and inference.

\paragraph{Benchmark evaluation.} As AIM is introduced as an additional component that's integrated into SIB \cite{Hu2020Empirical}, the gain in accuracy shows the importance of having a mixture of experts for fast adaptation. We also show the results for SIB with a linear layer (parameters equal the total parameters found in the AIM module) added before the classifier (SIB + Linear) to show that the gain in accuracy from AIM is not solely from the increase in parameters. From Table \ref{tab:few-shot}, we can see that AIM outperforms all existing few-shot classification methods by a noticeable margin. As only a single layer of AIM is explored, the coupling between AIM as found in RIMs \cite{goyal2019recurrent} is not considered here. We believe that further improvements can be attained if layers of AIM are stacked, with coupling between them considered.

\subsection{Continual Learning}\label{sec:continual}
\paragraph{Datasets.}
Omniglot \cite{lake2015human} has over 1,623 characters from 50 diferent alphabets, where each character has 20 hand-written images of size $28\times28$. The dataset is split into 963 classes for meta-training and 660 classes for meta-testing. In each trajectory, 15 images are used for training and 5 images for testing in both meta-training and meta-testing. 
CIFAR-100 \cite{krizhevsky2009learning} is composed of 60,000 images of size $32\times 32$ distributed uniformly over 100 classes, i.e.\ 500 train images and 100 test images for each class. 70 classes are used for meta-training and 30 classes are used for meta-testing. 
MiniImageNet \cite{vinyals2016matching} has 64 training classes and 20 testing classes with images of size $84\times84$. Each class has 600 images with 540 for training and 60 for testing. 30 training images are sampled for each class. In each trajectory of CIFAR-100 and MiniImageNet, we sample 30 train images for training all test images for testing for both meta-training and meta-testing.

\paragraph{Network architecture.}
We adopt the model from OML \cite{javed2019meta} and ANML \cite{beaulieu2020learning} with a slight modification for our experiments. For OML, the feature extractor $f_{\boldsymbol{\psi}}$ is a 6-layer convolutional network with 112 channels and the classifier $\boldsymbol{\phi}$ is a single linear layer with AIM $\mathcal{A}_{\boldsymbol{W}}$ in between $f_{\boldsymbol{\psi}}$ and $\boldsymbol{\phi}$. For ANML, both neuromodulatory network $f_{\boldsymbol{\psi}^{\mathrm{NM}}}$ and prediction network $f_{\boldsymbol{\psi}^{\mathrm{P}}}$ have a 3-layer convolutional network and $\boldsymbol{\phi}$ is a single linear layer with AIM placed after $f_{\boldsymbol{\psi}^{\mathrm{NM}}}$ and $f_{\boldsymbol{\psi}^{\mathrm{P}}}$. $f_{\boldsymbol{\psi}^{\mathrm{NM}}}$ has 112 channels while $f_{\boldsymbol{\psi}^{\mathrm{P}}}$ has 256 channels. For CIFAR-100 and MiniImageNet, an additional linear layer is placed before AIM for dimension reduction. The hidden state $\mathbf{h}_m\in \mathbb{R}^{128}$. $W^{\mathrm{K}}\in \mathbb{R}^{\mathrm{dim}(\hat{\mathbf{z}}) \times 128}$ and $W^{\mathrm{Q}}_m\in\mathbb{R}^{128 \times 128}$ maps theirs corresponding inputs to $\mathbb{R}^{128}$.


\paragraph{Training details.}
We use $M=64$ mechanisms in our system and top $K=10$ mechanisms are selected during inference with induced stochasticity by having $l=2$ during training. We follow the $1^\text{st}$-order MAML strategy in \cite{javed2019meta,beaulieu2020learning}. We use a batch size of 1 for 20,000 steps with step size of $\nu^{\mathrm{out}} = 1\times10^{-3}$ for the outer-loop (slow weights) and $\nu^{\mathrm{in}} = 1\times10^{-2}$ for the inner-loop (fast weights). A complete meta-training of AIM using OML or ANML on Omniglot, CIFAR-100 and MiniImageNet takes less than 2 hours, 3 hours and 6 hours respectively.

\subsubsection{Qualitative Study: Activation of AIM}
Following the settings in few-shot learning, activations of AIM when applied to OML are shown in Figure \ref{fig:oml_aim}. The activations are similar to what we observed in few-shot learning, i.e.\ a set of common mechanisms for all classes and another set for mechanisms that are sparsely activated.

\subsubsection{Quantitative Study}
To evaluate the capability of AIM to continually learn new concepts and mitigating catastrophic forgetting, we show the results of meta-test training and testing in Figure \ref{fig:benchmark_continual}. To demonstrate that the accuracy gain using AIM is not due to the increase in parameters, \textit{baseline} is plotted and is defined as the swapping of AIM with a linear layer containing the same amount of parameters as AIM added to OML. Samples of new classes are continuously fed without replacement, and samples of old classes are not stored. Prior works use the results from meta-test training as a measure of forgetting and meta-test testing to measure both forgetting and generalization error. We argue that memorizing features that doesn't transfer well to the testing set is not a good measure of forgetting. Results show that through the application of AIM, the difference between train and test accuracy is marginal, i.e.\ small generalization error, demonstrating that AIM is not only useful for the adaptation to new knowledge and mitigation of catastrophic forgetting, it also plays an important role in the learning of concepts that are generalizable to the test set. Consistent improvement in accuracy is observed when AIM is applied to existing continual learning frameworks. The only exception is the application of AIM to ANML trained on Omniglot, which could be remedied through a better selection of hyperparamters.

\section{Conclusion}
We have shown that AIM as a mixture of experts is an important building block for modeling higher-order concepts, translating to the capability of fast adaptation and mitigation of catastrophic forgetting. Through the sparse modeling of higher-order concepts, substantial improvement over prior arts can be seen for both few-shot and continual learning. It would be interesting to see the extension of AIM to multiple layers for hierarchical modeling of higher-order concepts.

\clearpage

\section*{Acknowledgement}
This project is supported by MOST under code 107-2221-E-009 -125 -MY3. Eugene Lee is partially supported by Novatek Ph.D.\ Fellowship Award. The authors are grateful for the suggestions provided by Dr.\ Eugene Wong from University of California in Berkeley and Dr.\ Jian-Ming Ho from Academia Sinica of Taiwan.

{\small
\bibliographystyle{ieee_fullname}
\bibliography{egpaper_final}
}

\clearpage

\appendix

\section{Few-Shot Learning}
\subsection{Observation of Attention Weight Change over Training Epochs}

\begin{figure*}[!htbp]
\centering
  	\subfloat[CIFAR-FS, 1-shot.]{\includegraphics[width=0.5\textwidth]{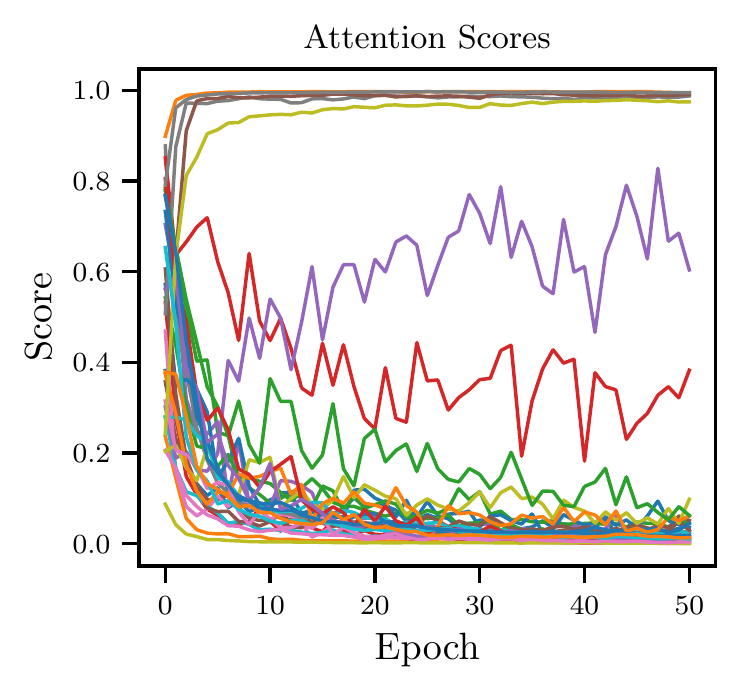}\label{fig:latency_vgg11_c10}}
	\hfill
	\subfloat[CIFAR-FS, 5-shot.]{\includegraphics[width=0.5\textwidth]{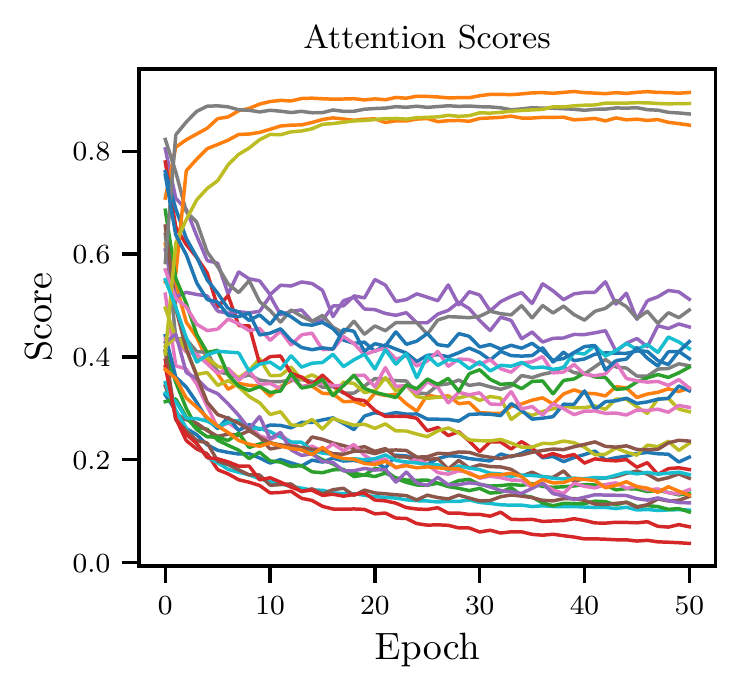}\label{fig:latency_vgg11_c100}}
	\vfill
  	\subfloat[MiniImageNet, 1-shot.]{\includegraphics[width=0.5\textwidth]{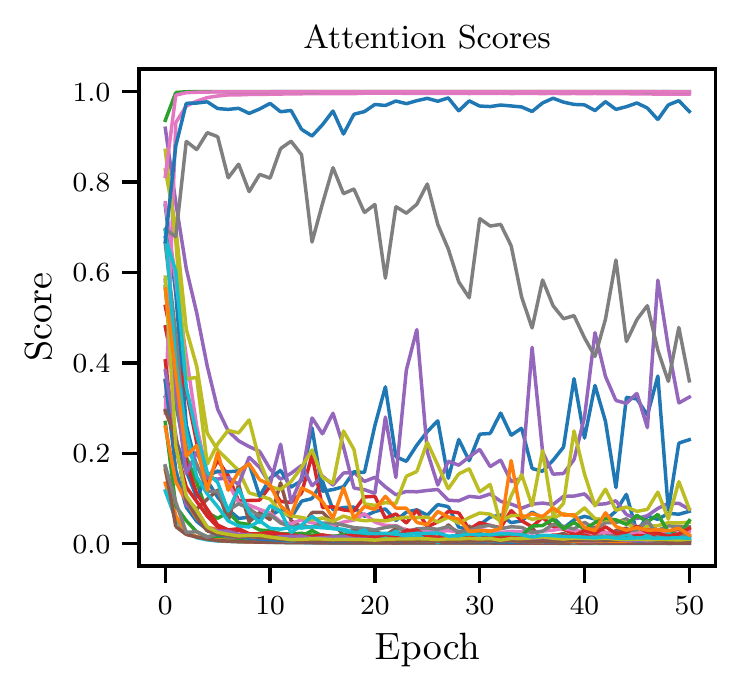}\label{fig:latency_vgg11_c10}}
	\hfill
	\subfloat[MiniImageNet, 5-shot.]{\includegraphics[width=0.5\textwidth]{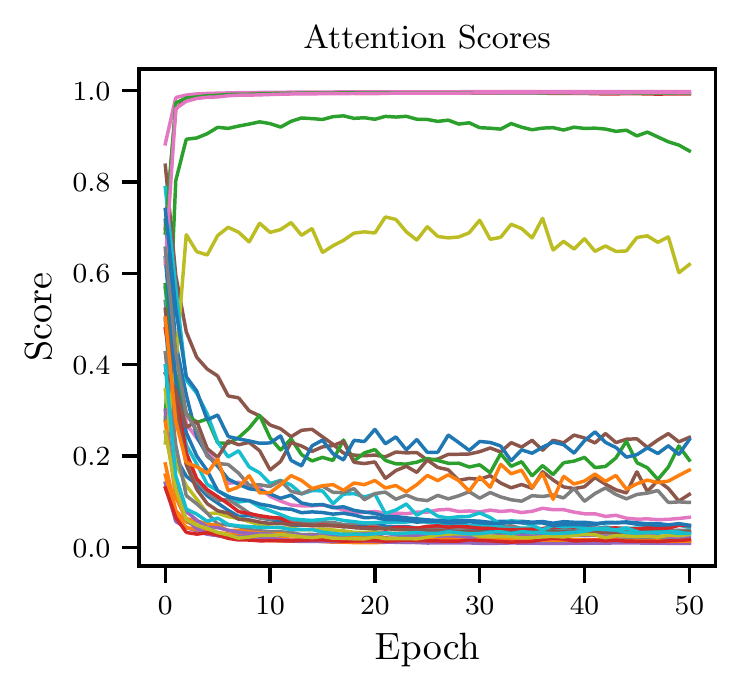}\label{fig:latency_vgg11_c100}}
\caption{Change of attention weight or score corresponding to the input dimension over the training epochs. Different datasets pairs with different amount of training samples (shots) are shown here, using Conv-4-64 as its backbone. Each line represent an independent mechanism.}
\label{fig:attn_weight_conv}
\end{figure*}

\begin{figure*}[!htbp]
\centering
  	\subfloat[CIFAR-FS, 1-shot.]{\includegraphics[width=0.5\textwidth]{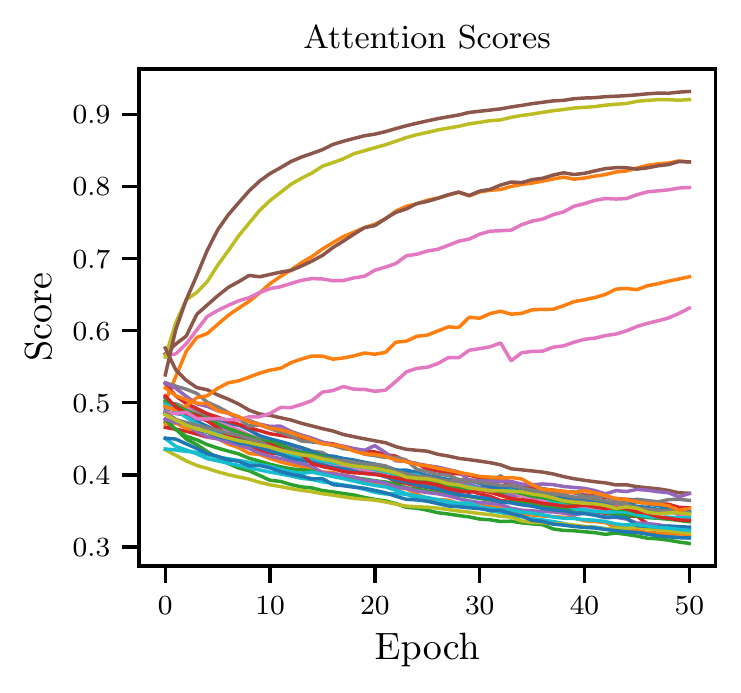}\label{fig:latency_vgg11_c10}}
	\hfill
	\subfloat[CIFAR-FS, 5-shot.]{\includegraphics[width=0.5\textwidth]{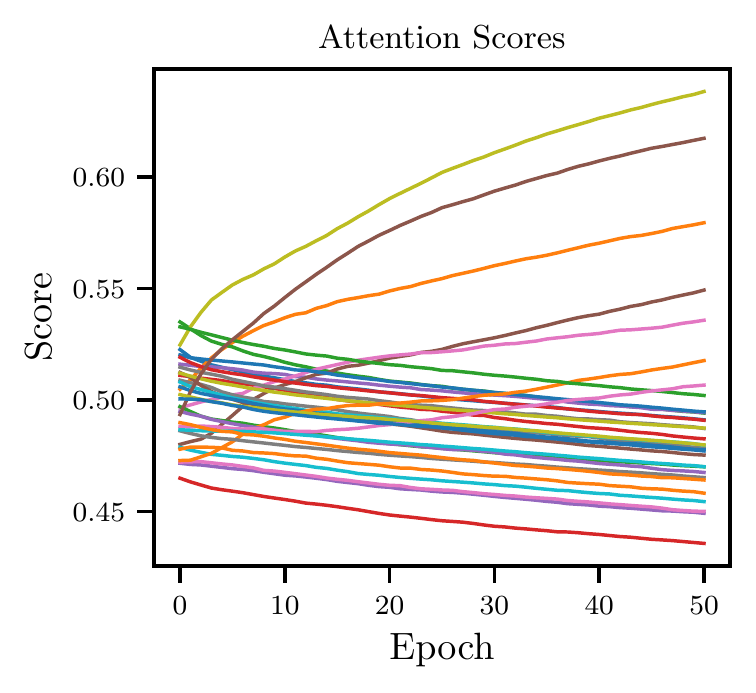}\label{fig:latency_vgg11_c100}}
	\vfill
  	\subfloat[MiniImageNet, 1-shot.]{\includegraphics[width=0.5\textwidth]{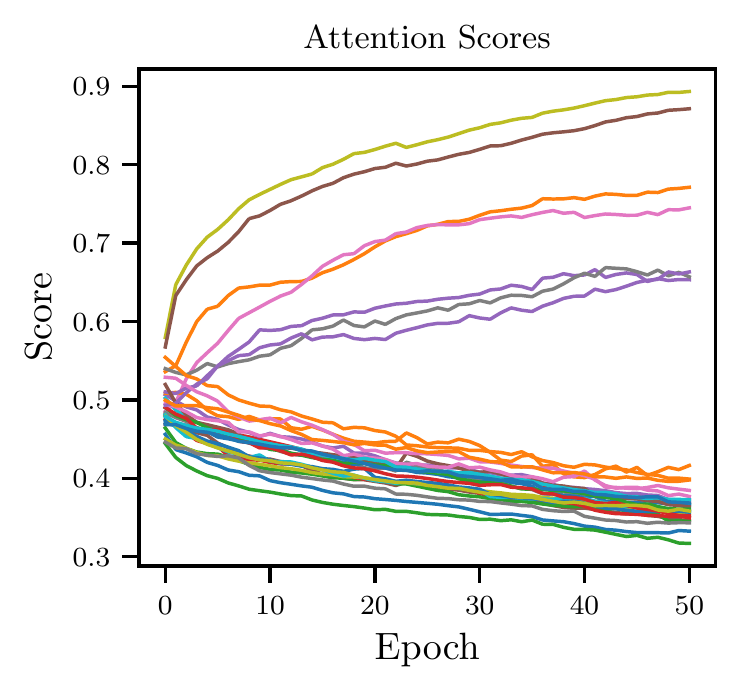}\label{fig:latency_vgg11_c10}}
	\hfill
	\subfloat[MiniImageNet, 5-shot.]{\includegraphics[width=0.5\textwidth]{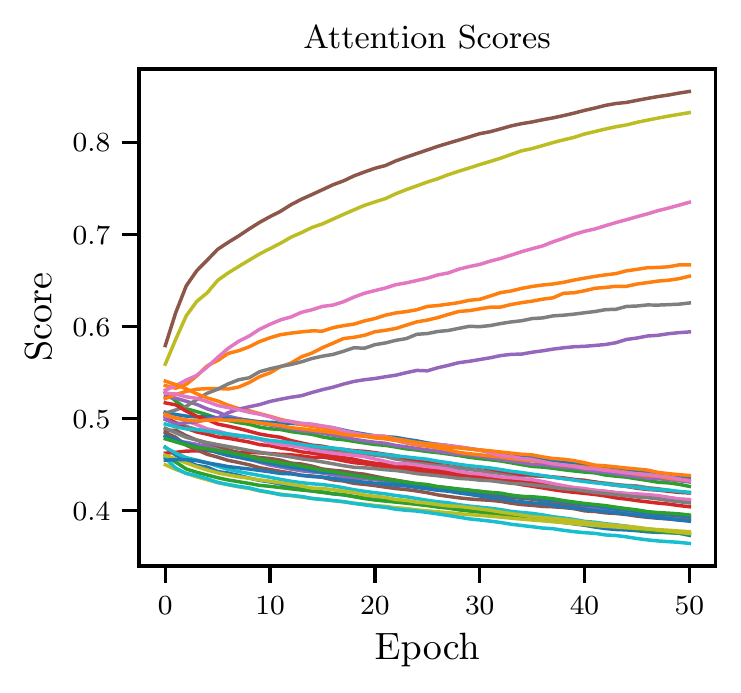}\label{fig:latency_vgg11_c100}}
\caption{Change of attention weight or score corresponding to the input dimension over the training epochs. Different datasets pairs with different amount of training samples (shots) are shown here, using WRN-28-10 as its backbone. Each line represent an independent mechanism. }
\label{fig:attn_weight_wrn}
\end{figure*}

In this section, we provide an empirical study on the change of the attention weight corresponding the the input dimension of the attention score, $\mathrm{softmax}\left(\frac{\langle \mathbf{h}_{m}W_{m}^{\mathrm{Q}}, \hat{\mathbf{z}} W^{\mathrm{K}} \rangle}{\sqrt{d}}\right)$, over the entire training epochs. This study provides two insights: 1. the dynamics of the activation of mechanisms based on the number of training samples (shots); 2. the distribution of active and inhibited mechanisms over the training iterations. To show the dynamical change in a 2D plot, we sample classes from the validation set and observe them for the entire training process. Plots that uses Conv-4-64 and WRN-28-10 as backbone is shown in Figure \ref{fig:attn_weight_conv} and Figure \ref{fig:attn_weight_wrn} respectively. From the plots, we can see that the activation of mechanisms are initially distributed uniformly followed by slow convergence to a sparse distribution over the training epochs, having only a few active mechanisms upon convergence. The active and inhibited attention weights are also clearly separated for all examples. Another observation is that having a larger number of training samples (shots), a smoother convergence for the activation weights across the training epochs is obtained. Smooth convergence is also obtained when a deeper backbone (WRN-28-10) is used when compared to a shallower one (Conv-4-64). This observation is intuitive as AIM is able to learn more efficiently when more samples or higher quality input features are provided, enabling the mechanisms to better model higher-order factorized information.

\paragraph{Competitive selection of mechanisms.} From the figures shown, a distinct gap between active and inhibited mechanisms can be clearly observed. This motivates the idea of basing the activation of mechanisms on its corresponding attention value (soft decision) instead of making a hard decision that selects a total of $K$ AIM on every inference. To demonstrate if basing the activation of AIM on the attention value would work, a simple experiment can be performed by allowing a mechanism to be active if its attention value is above 0.5 (similar to ReLU \cite{nair2010rectified}), or:
\begin{equation}
    \tilde{\mathbf{z}} = \hat{\mathbf{z}}\left(\sum_{m=1}^{M}w_{m}(\hat{\mathbf{z}})W_{m}^{\text{M}}\right), \label{eq:weighted}
\end{equation}
where $w_{m}(\hat{\mathbf{z}})$ is given as,
\begin{equation}
w_{m}(\hat{\mathbf{z}}) = \begin{cases} \widetilde{w}_{m}(\hat{\mathbf{z}}), & \text{if } m \in \{ m \mid \widetilde{w}_{m}(\hat{\mathbf{z}}) > 0.5 \mathrm{\ and }\\
& \qquad\qquad\qquad\qquad 1 \leq m \leq  M\}, \\
0, & \text{otherwise,} \end{cases}    \label{eq:topk}
\end{equation}
and $\widetilde{w}_{m}(\hat{\mathbf{z}})$ is defined as,
\begin{equation}
\widetilde{w}_{m}(\hat{\mathbf{z}}) = \mathrm{softmax}\left(\frac{\langle \mathbf{h}_{m}W_{m}^{\mathrm{Q}}, \hat{\mathbf{z}} W^{\mathrm{K}} \rangle}{\sqrt{d}}\right).\label{eq:softmax}
\end{equation}
To keep our experiments simple, we do not induce stochasticity in the original approach and fix $K=8$ to provide a fair comparison. We name the original method that keeps 8 mechanisms active as \textit{hard decision} and name the method in (\ref{eq:weighted}) -- (\ref{eq:softmax}) as \textit{soft decision}. Comparison between hard decision and soft decision is shown in Table \ref{tab:soft_hard}. From the results, we can see that when Conv-4-64 is used as backbone, higher accuracy is obtained when hard decision is used. The opposite can be observed when WRN-28-10 is used as backbone. We deduce that when extracted features are more reliable, i.e.\ through the use of deeper backbone or higher number of shots, the attention weights are of higher quality leading to clear distinction between relevant and less relevant mechanisms.

\begin{table*}[!htbp]
\scriptsize
  \caption{Results for the comparison of using either hard decision (proposed method; $K=8$ with $l=0$) or soft decision (\ref{eq:weighted}) -- (\ref{eq:softmax}) for the activation of mechanisms during inference. Average classification accuracies with 95\% confidence intervals on the test-set are shown.}
  \label{tab:soft_hard}
  \centering
  \begin{tabular}{cccccc}
    \toprule
    \multirow{2}{*}{Backbone} &  \multirow{2}{*}{Method} & \multicolumn{2}{c}{MiniImageNet, 5-Way} & \multicolumn{2}{c}{CIFAR-FS, 5-Way}                   \\
       &  & 1-shot     & 5-shot & 1-shot     & 5-shot \\
    \midrule
    \multirow{2}{*}{Conv-4-64} 
                & Hard decision & $\mathbf{61.90\pm0.56\%}$  & $\mathbf{74.55\pm0.38\%}$ & $\mathbf{70.80\pm0.61\%}$ & $\mathbf{80.50\pm0.40\%}$    \\
                & Soft decision & $61.74\pm0.57\%$ & $74.41\pm0.38\%$  & $70.18\pm0.61\%$ & $80.38\pm0.39\%$    \\
    \midrule
    \multirow{2}{*}{WRN-28-10} 
                & Hard decision & $\mathbf{71.03\pm0.57\%}$  & $82.30\pm0.33\%$ & $79.19\pm0.55\%$ & $87.04\pm0.36\%$    \\
                & Soft decision & $69.96\pm0.56\%$  & $\mathbf{82.37\pm0.33\%}$ & $\mathbf{80.14\pm0.55\%}$ & $\mathbf{87.41\pm0.35\%}$  \\
    \bottomrule
  \end{tabular}
\end{table*}

\subsection{Observation of Attention Weight for All Classes}
\begin{figure*}[!htbp]
\centering
  	\subfloat[First epoch.]{\includegraphics[width=0.5\textwidth]{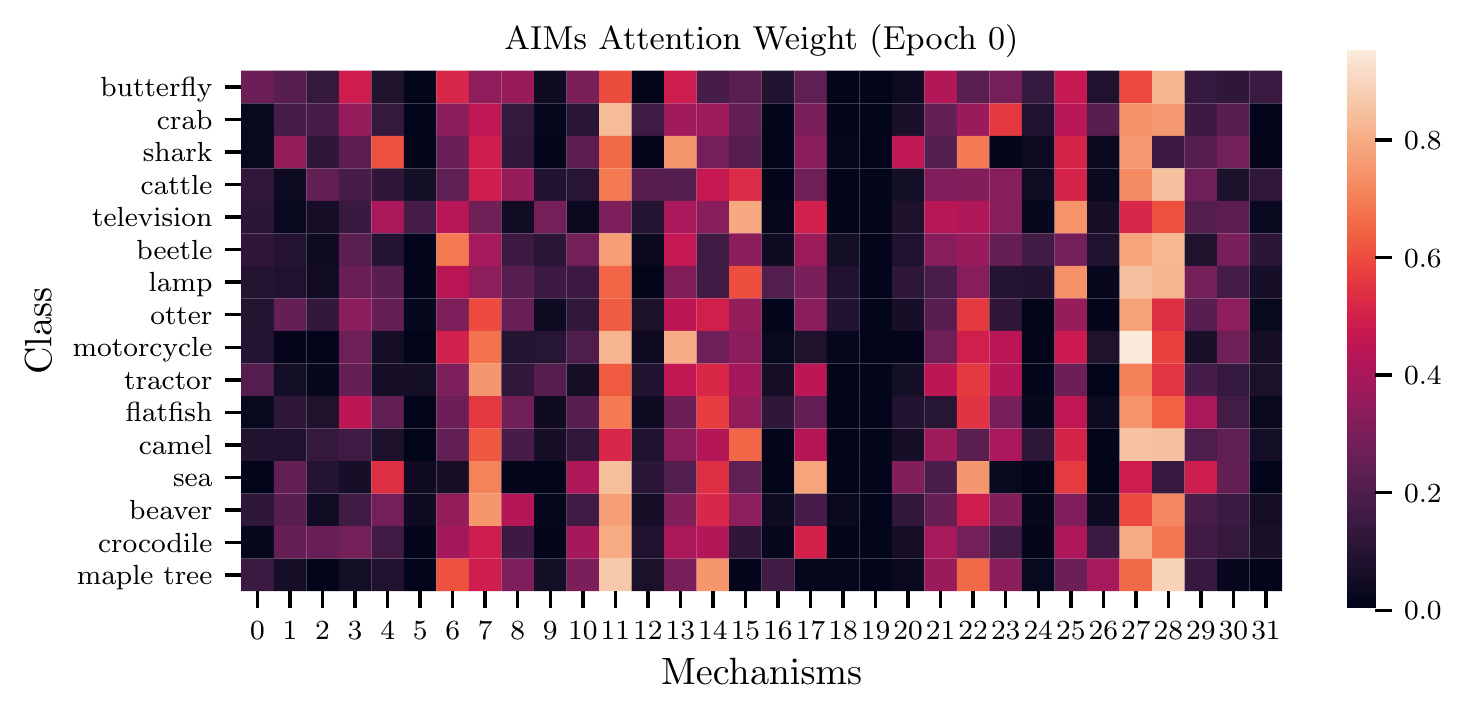}\label{fig:latency_vgg11_c10}}
	\hfill
	\subfloat[Last epoch.]{\includegraphics[width=0.5\textwidth]{images/mask-last_cifar_conv_1shot.pdf}\label{fig:latency_vgg11_c100}}
\caption{Activation of AIM from few-shot learning. Training on CIFAR-FS with 1-shot using Conv-4-64 as backbone. }
\label{fig:mask-cifarfs_1shot_conv}
\end{figure*}

\begin{figure*}[!htbp]
\centering
  	\subfloat[First epoch.]{\includegraphics[width=0.5\textwidth]{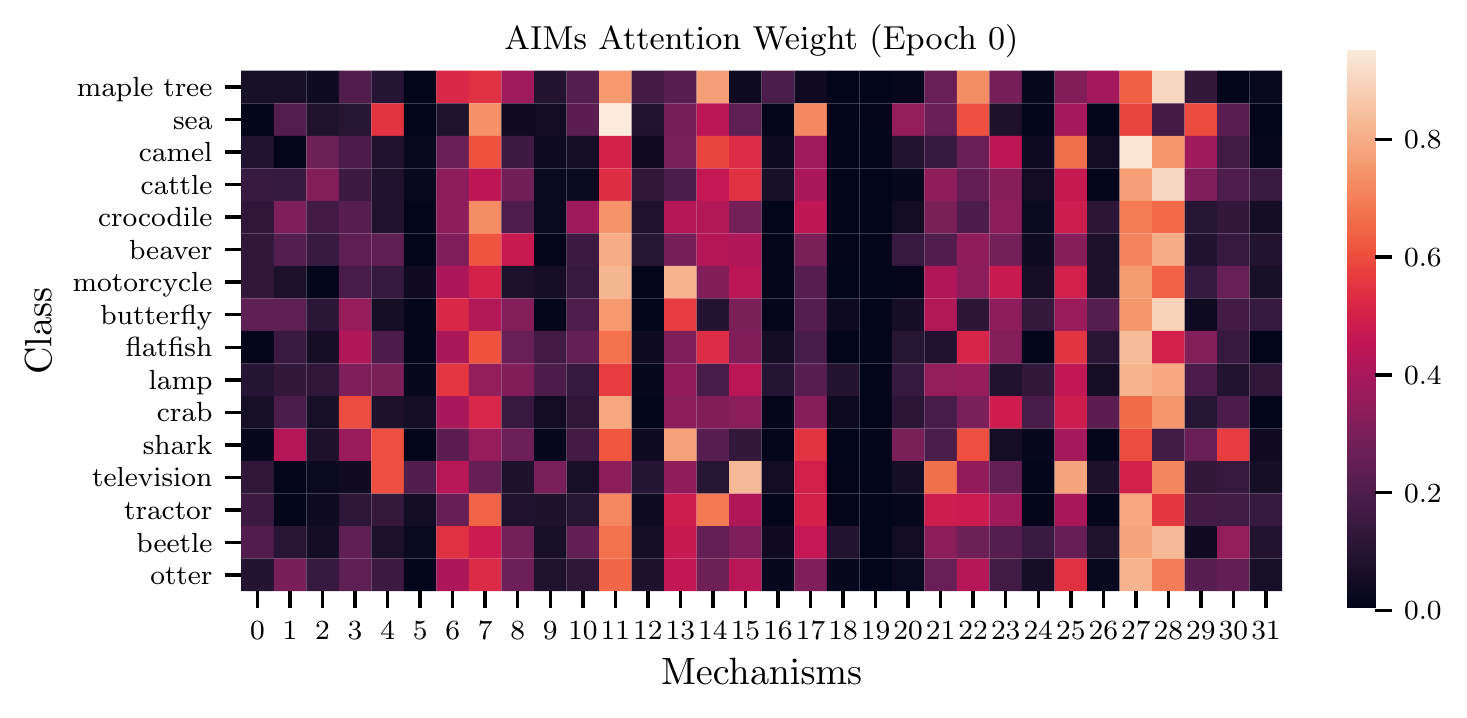}\label{fig:latency_vgg11_c10}}
	\hfill
	\subfloat[Last epoch.]{\includegraphics[width=0.5\textwidth]{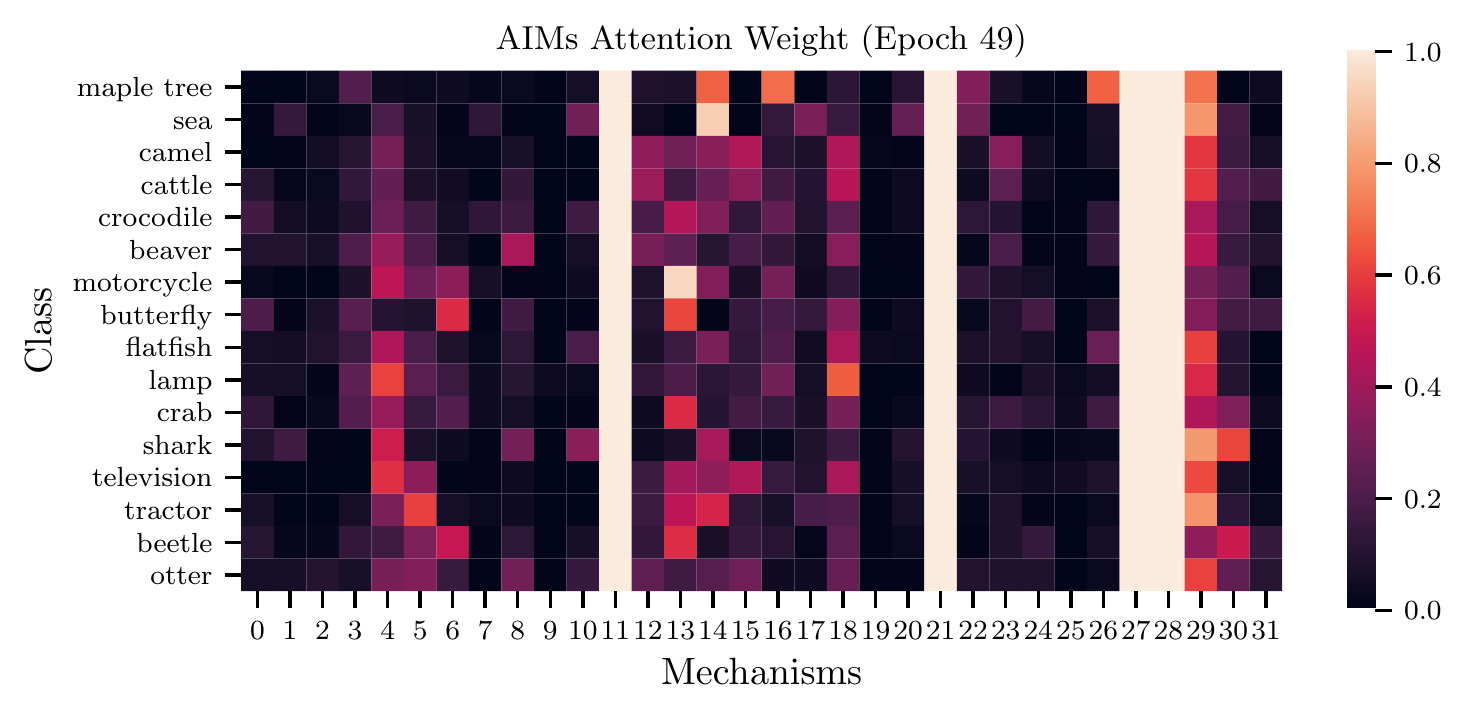}\label{fig:latency_vgg11_c100}}
\caption{Activation of AIM from few-shot learning. Training on CIFAR-FS with 5-shot using Conv-4-64 as backbone. }
\label{fig:mask-cifarfs_5shot_conv}
\end{figure*}

\begin{figure*}[!htbp]
\centering
  	\subfloat[First epoch.]{\includegraphics[width=0.5\textwidth]{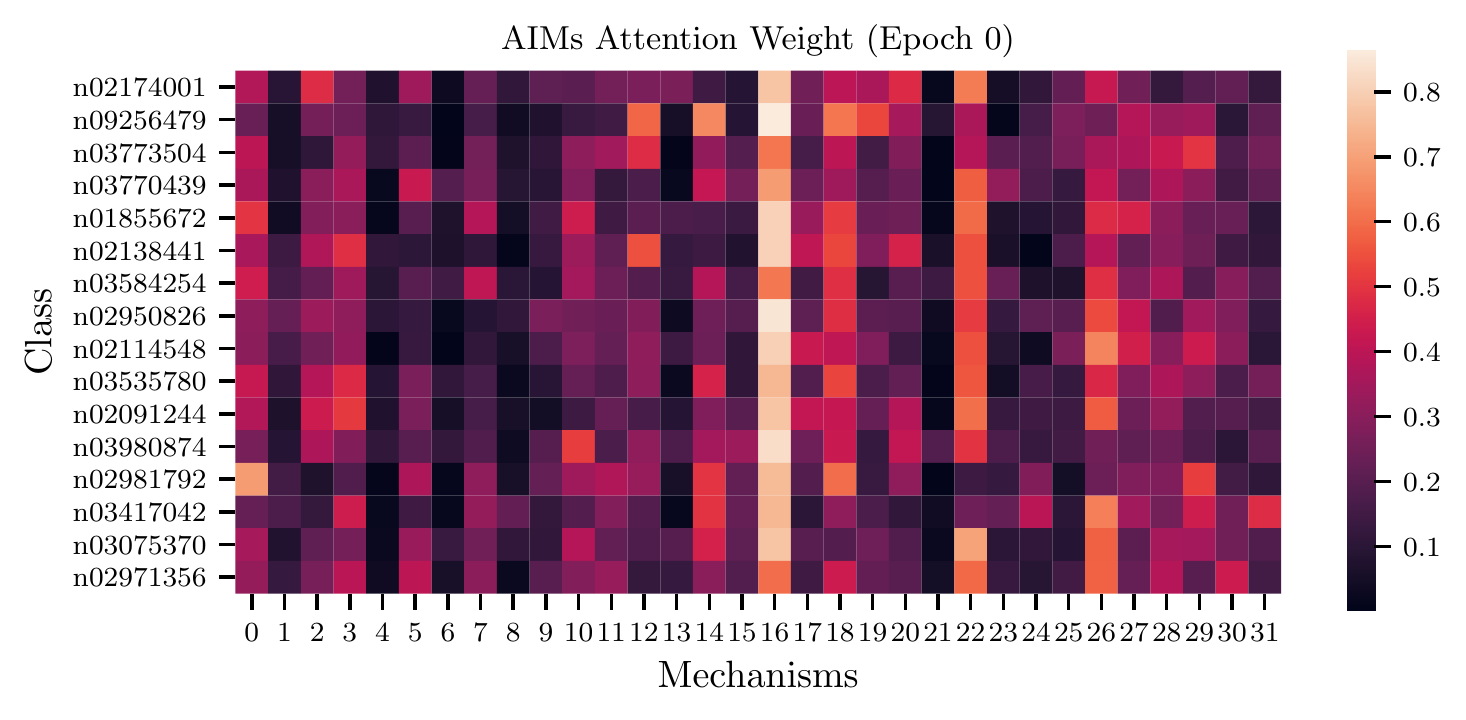}\label{fig:latency_vgg11_c10}}
	\hfill
	\subfloat[Last epoch.]{\includegraphics[width=0.5\textwidth]{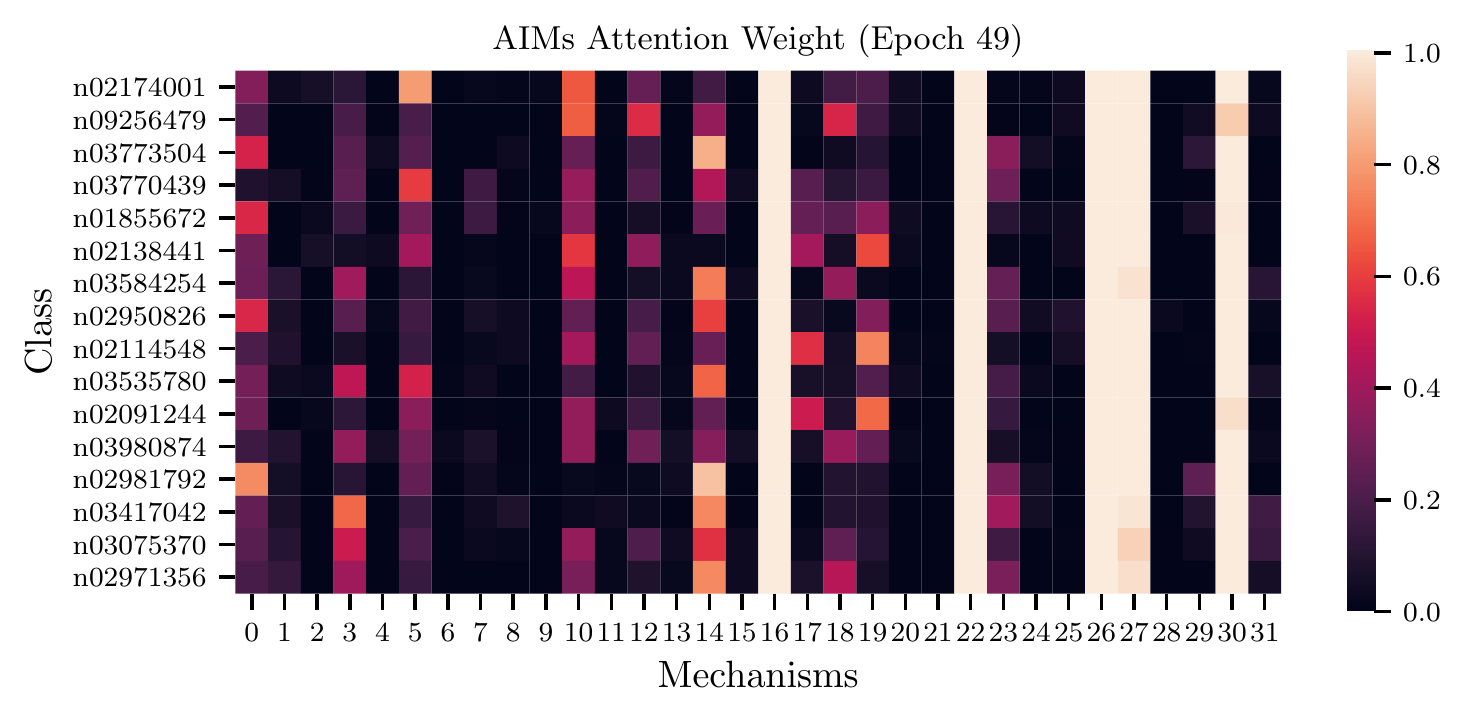}\label{fig:latency_vgg11_c100}}
\caption{Activation of AIM from few-shot learning. Training on MiniImageNet with 1-shot using Conv-4-64 as backbone. }
\label{fig:mask-mini_1shot_conv}
\end{figure*}

\begin{figure*}[!htbp]
\centering
  	\subfloat[First epoch.]{\includegraphics[width=0.5\textwidth]{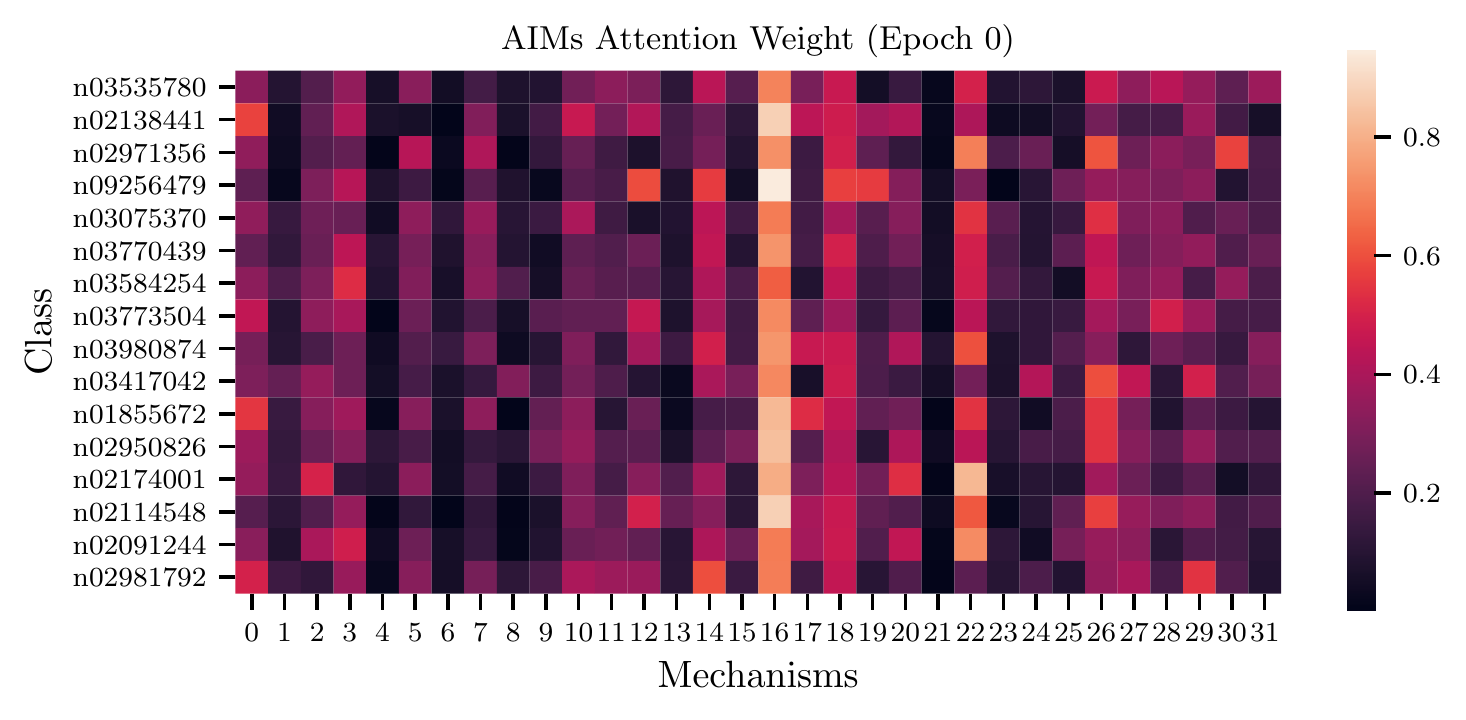}\label{fig:latency_vgg11_c10}}
	\hfill
	\subfloat[Last epoch.]{\includegraphics[width=0.5\textwidth]{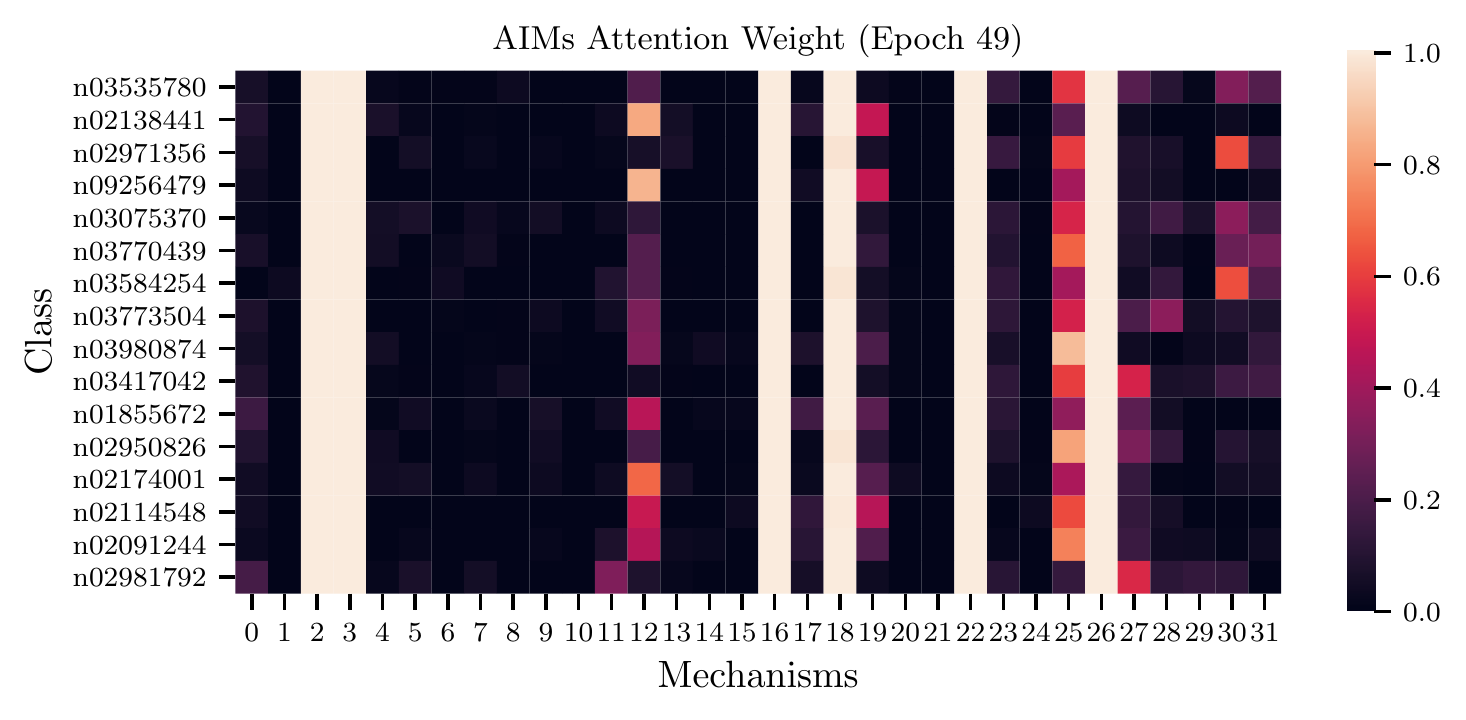}\label{fig:latency_vgg11_c100}}
\caption{Activation of AIM from few-shot learning. Training on MiniImageNet with 5-shot using Conv-4-64 as backbone. }
\label{fig:mask-mini_5shot_conv}
\end{figure*}

\begin{figure*}[!htbp]
\centering
  	\subfloat[First epoch.]{\includegraphics[width=0.5\textwidth]{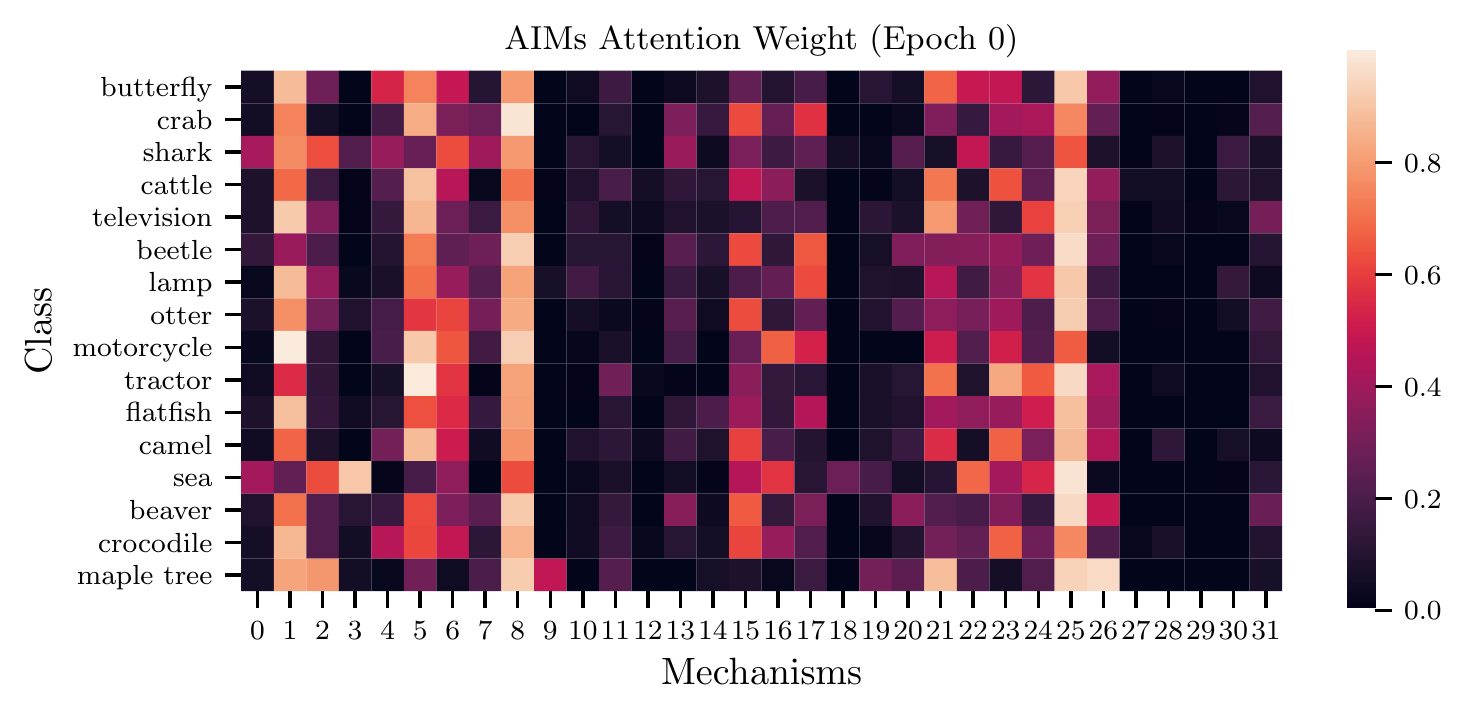}\label{fig:latency_vgg11_c10}}
	\hfill
	\subfloat[Last epoch.]{\includegraphics[width=0.5\textwidth]{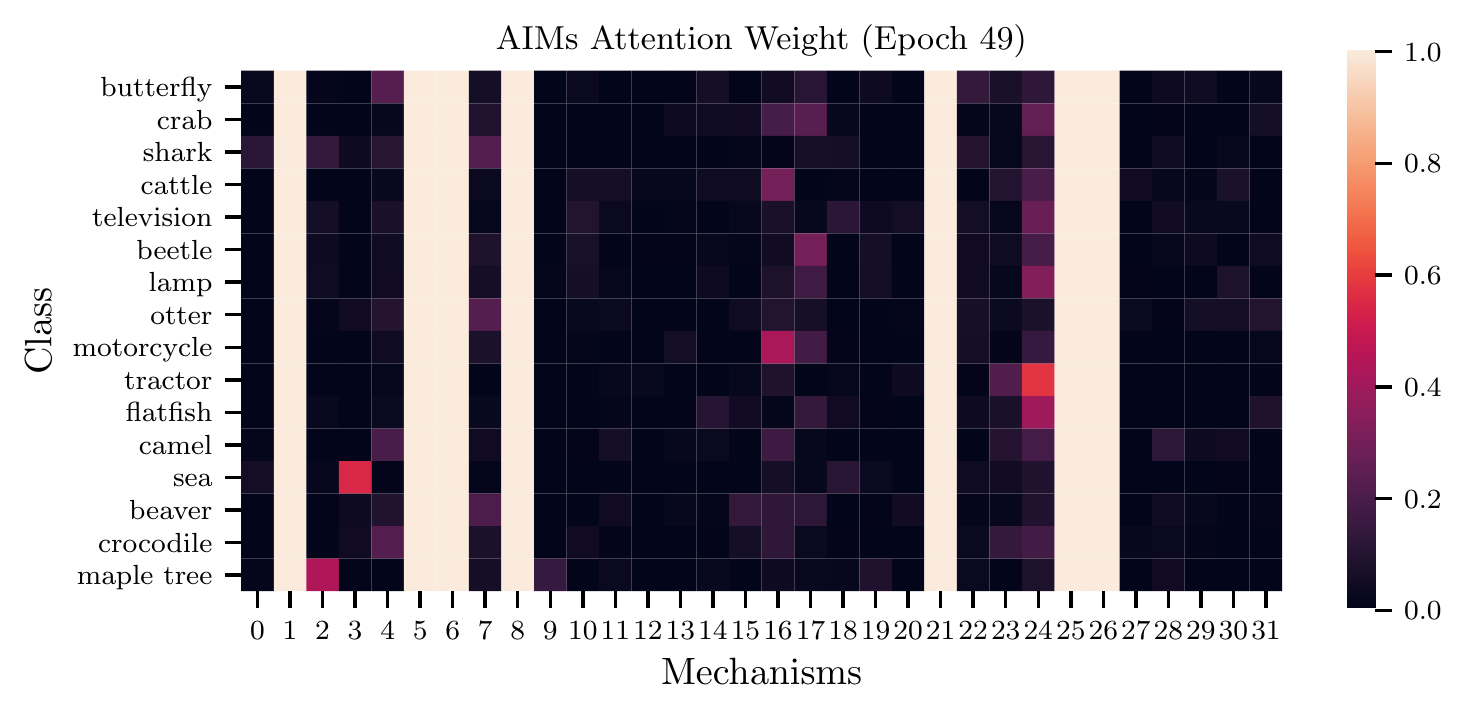}\label{fig:latency_vgg11_c100}}
\caption{Activation of AIM from few-shot learning. Training on CIFAR-FS with 1-shot using WRN-28-10 as backbone. }
\label{fig:mask-cifar_1shot_wrn}
\end{figure*}

\begin{figure*}[!htbp]
\centering
  	\subfloat[First epoch.]{\includegraphics[width=0.5\textwidth]{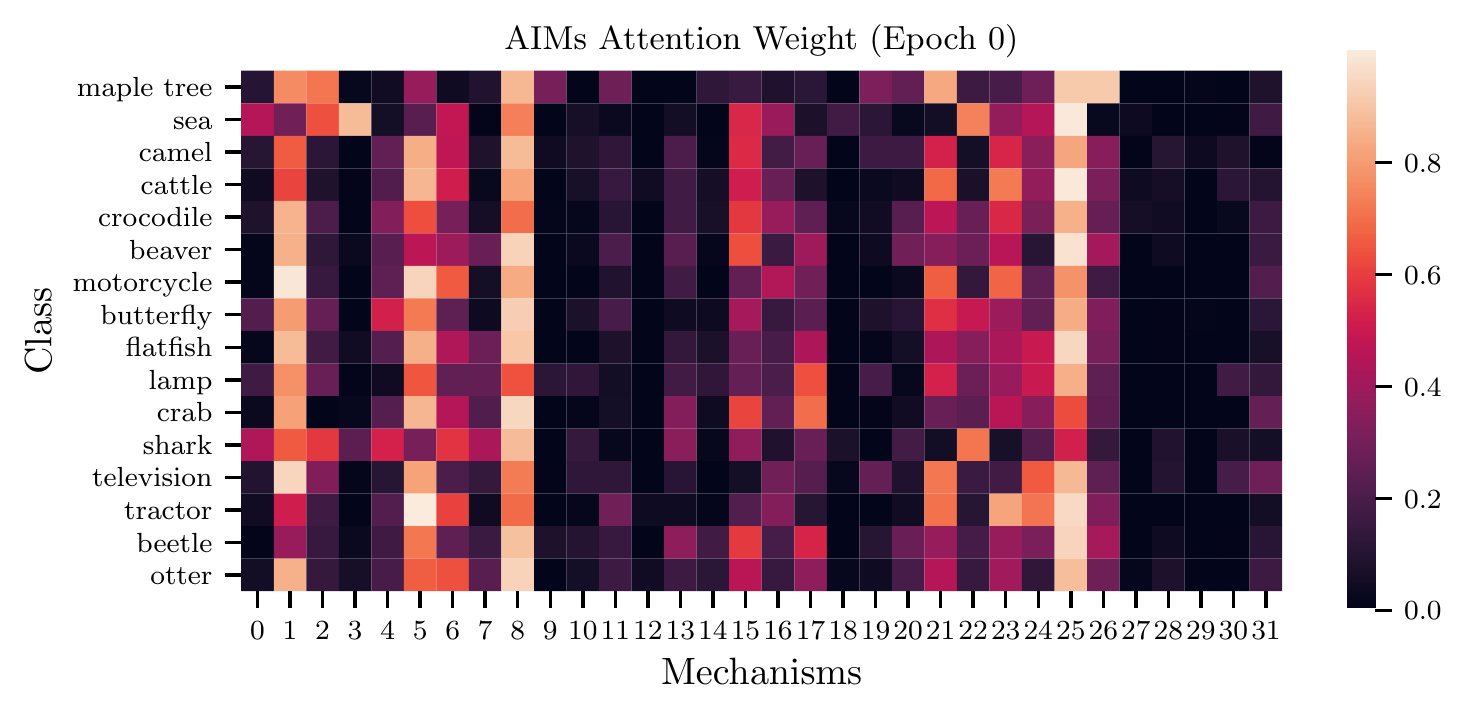}\label{fig:latency_vgg11_c10}}
	\hfill
	\subfloat[Last epoch.]{\includegraphics[width=0.5\textwidth]{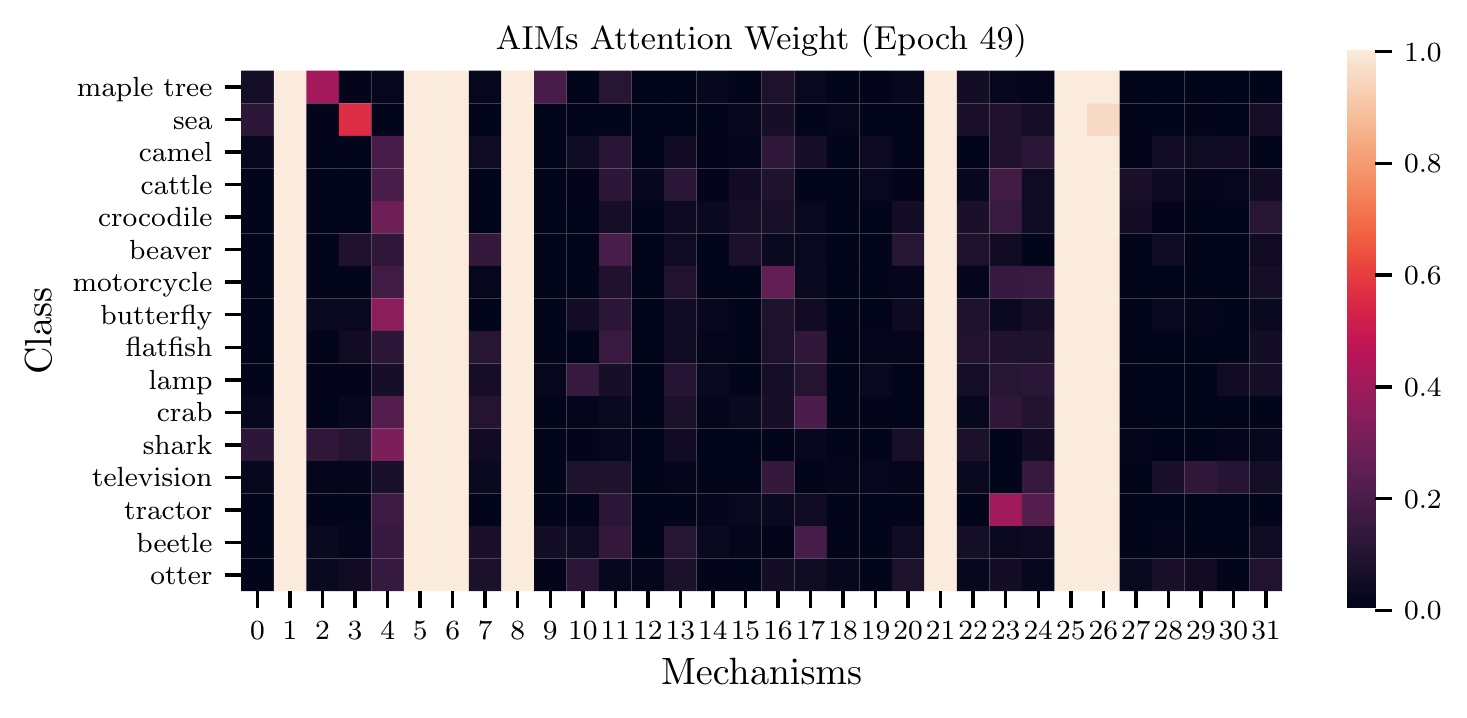}\label{fig:latency_vgg11_c100}}
\caption{Activation of AIM from few-shot learning. Training on CIFAR-FS with 5-shot using WRN-28-10 as backbone. }
\label{fig:mask-cifar_5shot_wrn}
\end{figure*}

\begin{figure*}[!htbp]
\centering
  	\subfloat[First epoch.]{\includegraphics[width=0.5\textwidth]{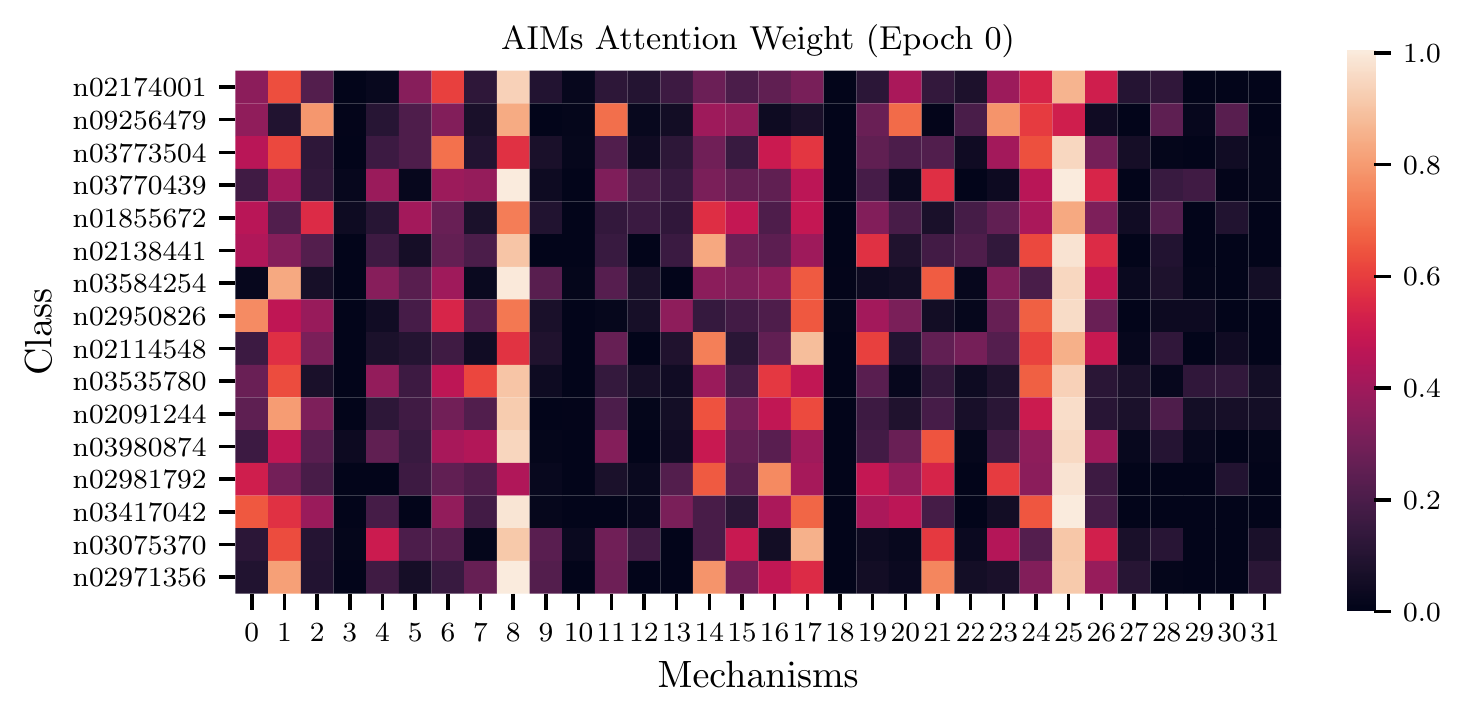}\label{fig:latency_vgg11_c10}}
	\hfill
	\subfloat[Last epoch.]{\includegraphics[width=0.5\textwidth]{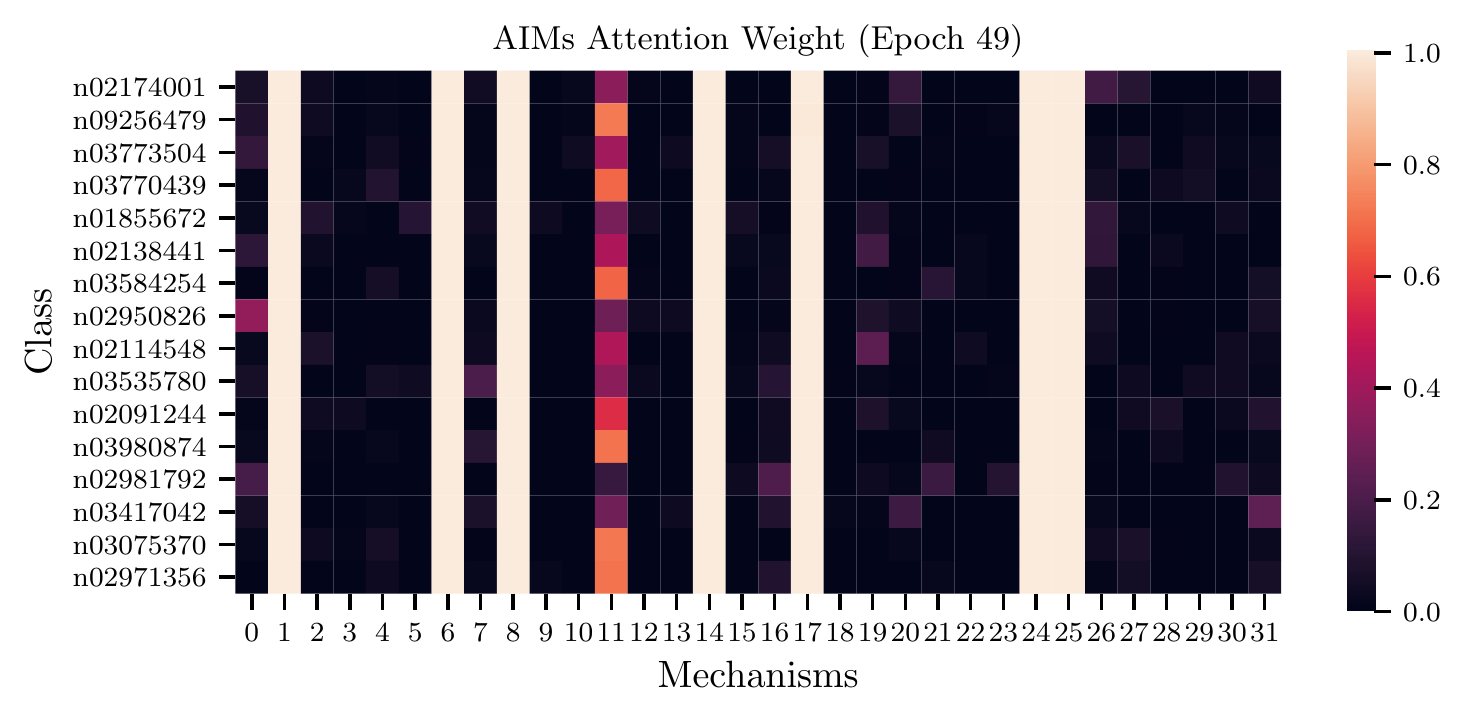}\label{fig:latency_vgg11_c100}}
\caption{Activation of AIM from few-shot learning. Training on MiniImageNet with 1-shot using WRN-28-10 as backbone. }
\label{fig:mask-mini_1shot_wrn}
\end{figure*}

\begin{figure*}[!htbp]
\centering
  	\subfloat[First epoch.]{\includegraphics[width=0.5\textwidth]{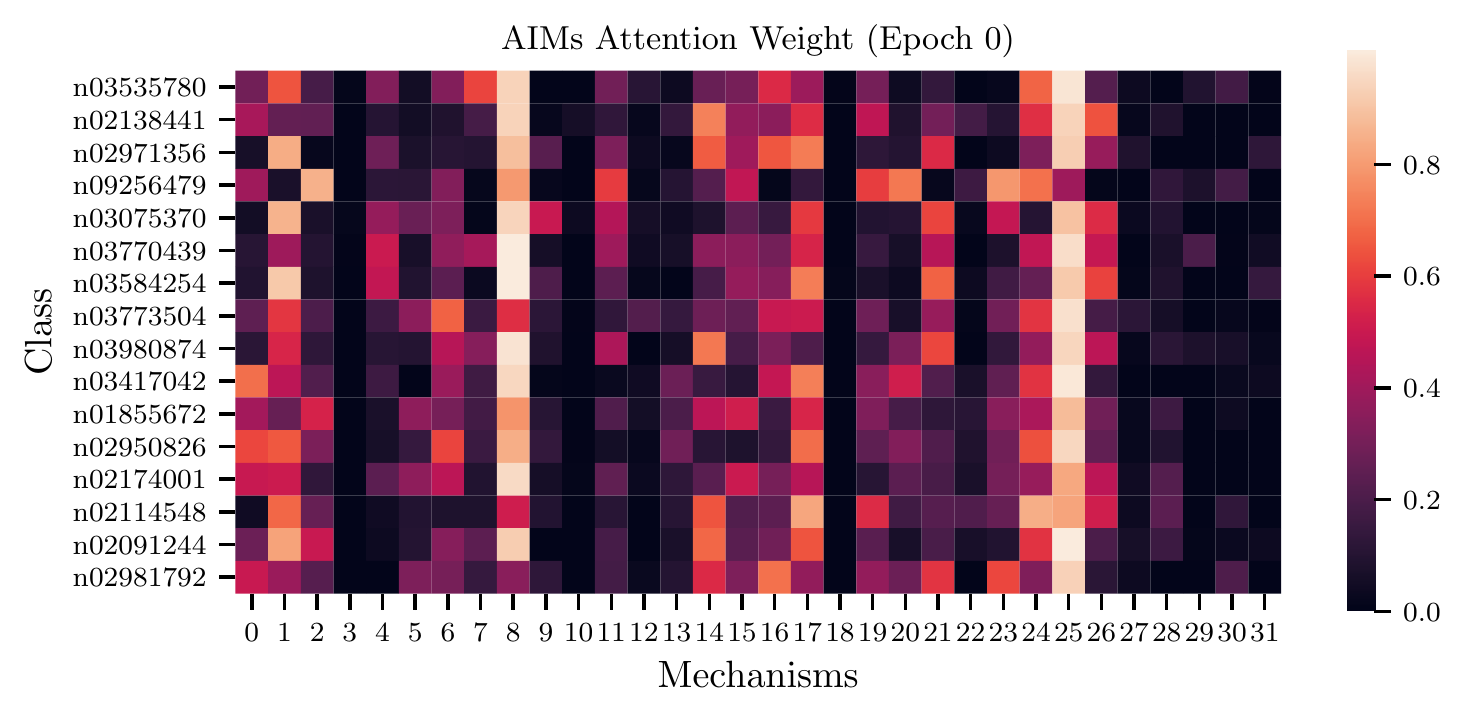}\label{fig:latency_vgg11_c10}}
	\hfill
	\subfloat[Last epoch.]{\includegraphics[width=0.5\textwidth]{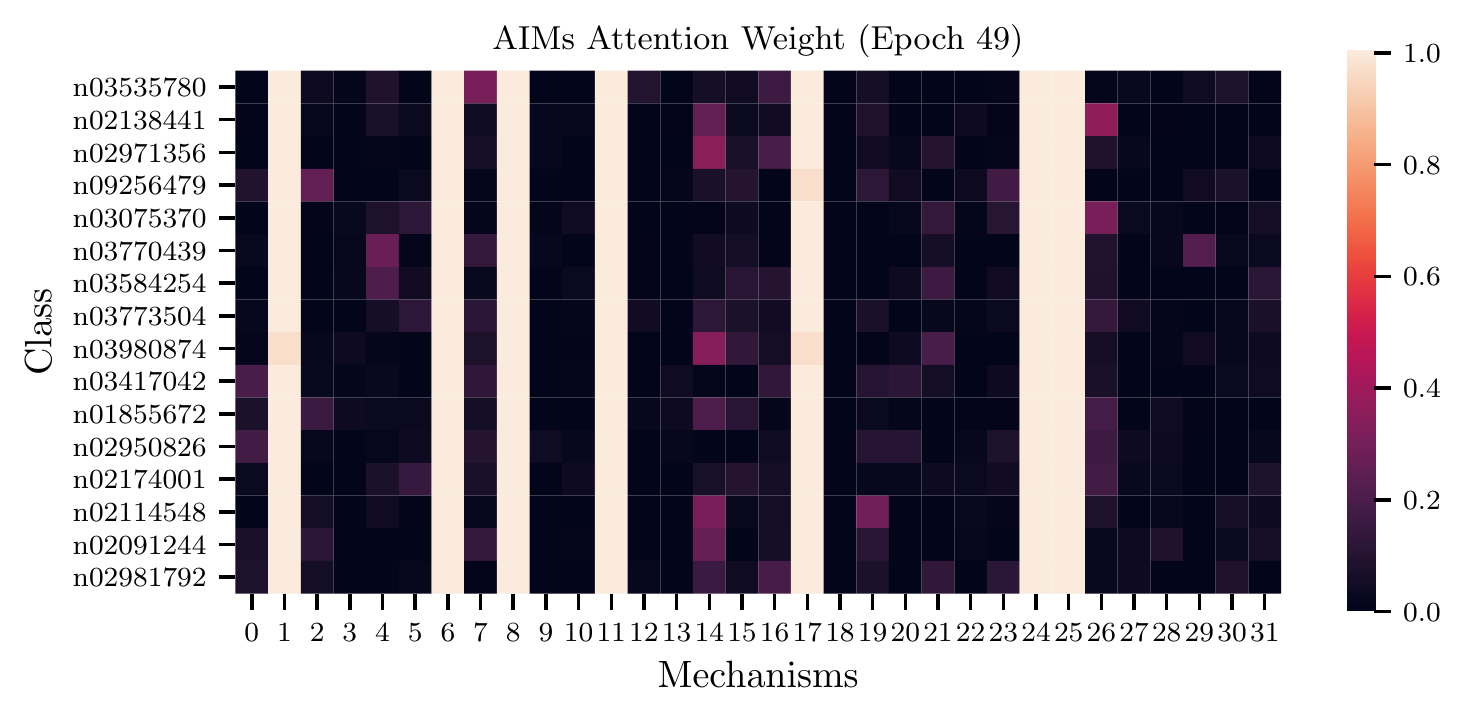}\label{fig:latency_vgg11_c100}}
\caption{Activation of AIM from few-shot learning. Training on MiniImageNet with 5-shot using WRN-28-10 as backbone. }
\label{fig:mask-mini_5shot_wrn}
\end{figure*}

Different from the previous section, we show the mask instead of the attention weights here. The masks have a value of 1 for active mechanisms and 0 for inhibited mechanisms having the competitive selection based on the attention weights; for all experiments, $K=8$ mechanisms will be active during both training and inference. We set $l=2$ to induce stochasticity during training. Instead of sampling a single sample from each class, we take the average of the masks of each class accumulated across the entire validation set. We show heatmaps covering all classes and all 32 AIMs mechanisms on the first and final training epochs. Results that use Conv-4-64 as backbone using \#shots-dataset pair of 1-shot-CIFAR-FS, 5-shot-CIFAR-FS, 1-shot-MiniImageNet and 5-shot-MiniImageNet are shown in Figure \ref{fig:mask-cifarfs_1shot_conv}, Figure \ref{fig:mask-cifarfs_5shot_conv}, Figure \ref{fig:mask-mini_1shot_conv} and Figure \ref{fig:mask-mini_5shot_conv} respectively. Results that use WRN-28-10 as backbone using \#shots-dataset pair of 1-shot-CIFAR-FS, 5-shot-CIFAR-FS, 1-shot-MiniImageNet and 5-shot-MiniImageNet are shown in Figure \ref{fig:mask-cifar_1shot_wrn}, Figure \ref{fig:mask-cifar_5shot_wrn}, Figure \ref{fig:mask-mini_1shot_wrn} and Figure \ref{fig:mask-mini_5shot_wrn} respectively. By observing the heatmaps, we can see that the activation of mechanisms in the first epoch are uniformly distributed whereas in the final epoch, only a few set of mechanisms that are jointly used between classes accompanied by a sparse set of mechanisms that are invariant among samples from the same class. This observation meets our expectation of learning a set of experts that are each responsible for certain task. To give a better illustration on the transition of the heatmaps over the training epochs, we have attached several \textbf{.mp4} files that follows the naming convention of \underline{mask-\textit{DATASET}\_\textit{MODEL}\_\textit{\#}shot} (italicized as wildcard strings) along with the supplementary materials.

\subsection{Manipulating the Stochastic Sampling and Active Mechanisms Count}
In this section, we show tabulated results of the manipulation of stochastic sampling and active mechanisms count as found in the main paper. The plots in the main paper show zero mean-ed results whereas the actual accuracy is reported in Table \ref{tab:sto_count} and Table \ref{tab:active_mech} for the maniputation of stochastic sampling count and active mechanisms count respectively. By looking at the tabulated results, we can say that the introduction of some stochasticity on the competitive selection of mechanisms during training is beneficial for the overall performance.

\section{Continual Learning}
\subsection{Quantitative Analysis}
We show the plots from the main paper comparing different continual learning methods in Figure \ref{fig:benchmark_continual} with the accompanied tabulated data in Table \ref{tab:continualz}. Baseline shown correspond to the swapping of AIM layer with a single linear layer with number of parameters close to the originally introduced AIM layer to demonstrate that the increase in accuracy is not from over-parameterization. From the results, we can observe that with the addition of AIM as a module for continual learning, consistent improvement in accuracy can be obtained. It is also shown that the gain in accuracy does not result from the increase in parameters as shown by the accuracy attained using the baseline method.

\subsection{Activation of AIM}
Similar to the analysis done for the activations of mechanisms for few-shot, we show the activations of mechanisms when AIM is used for continual learning. We apply AIM to both OML \cite{javed2019meta} and ANML \cite{beaulieu2020learning} with activation heatmaps when trained on Omniglot \cite{lake2015human}, CIFAR-100 \cite{krizhevsky2009learning} and MiniImageNet \cite{vinyals2016matching} in Figure \ref{fig:omniglot}, Figure \ref{fig:cifar100} and Figure \ref{fig:miniimagenet} respectively. We can observe that for Omniglot, the activation of mechanisms are sparsely distributed when compared to the activations obtained using CIFAR-100 and MiniImageNet. We conjecture that this is due to the simplicity of extracted representations, resulting in simpler higher-order modeling by the mechanisms. For natural images like CIFAR-100 and MiniImageNet, the features are not as distinct as the alphabets found in Omniglot, hence higher-order modeling of representations isn't as sparsely distributed. The sparsely distributed activations found in Omniglot result in distinctive increase in accuracy when compared to other datasets as shown in Table \ref{tab:continualz}, e.g.\ at a trajectory containing 600 classes, the relative increase in accuracy when AIM is applied to OML is 20.70\%. Even when MiniImageNet is used, distinctive increment in accuracy can also be observed, e.g.\ 19.40\% relative increment in accuracy when AIM is applied to ANML, which be believe is due to the richness of information embedded in the latent representation resulting from the larger image size of $84\times84$. We believe that larger gain in accuracy can be attained through the introduction of a better feature extractor, i.e.\ an alternative to convolutional layers.

\begin{figure*}[!htbp]
\centering
  	\subfloat[OML+AIM]{\includegraphics[width=0.5\textwidth]{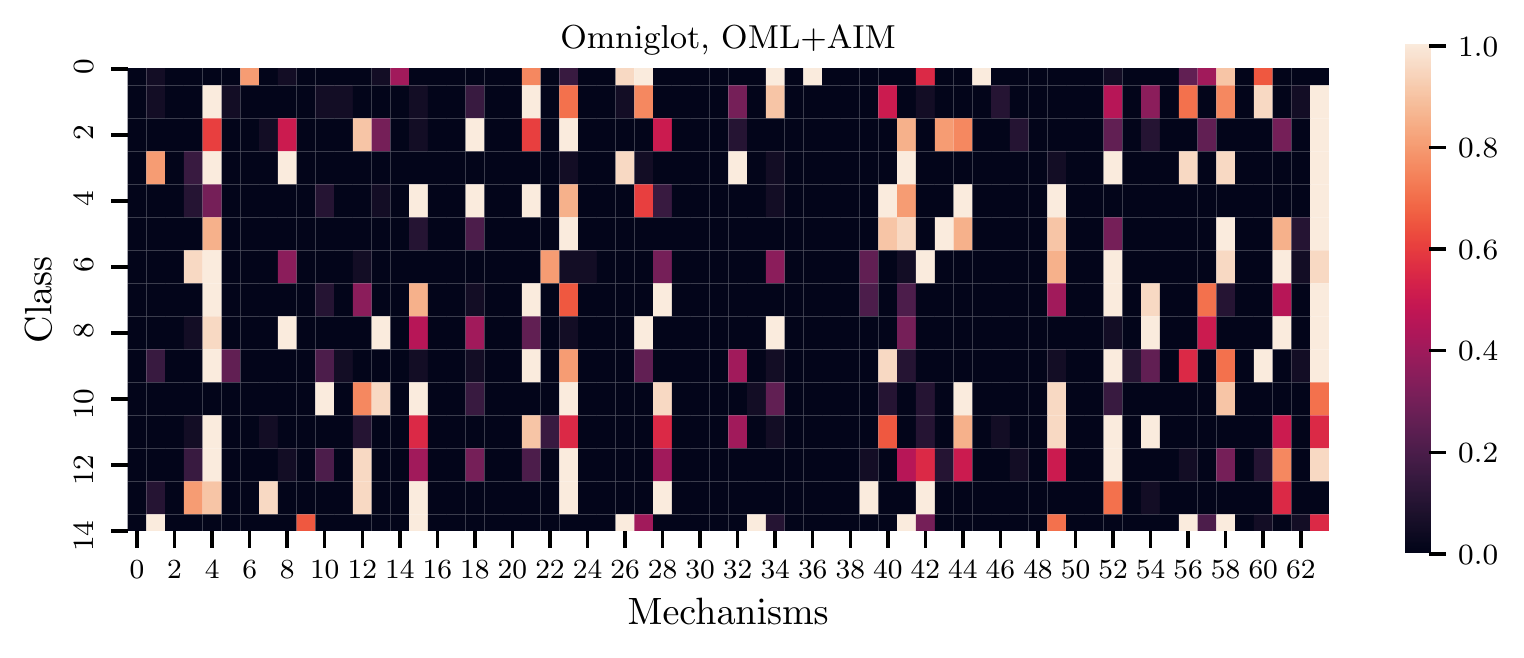}\label{fig:heat_omniglot_oml}}
	\hfill
	\subfloat[ANML+AIM]{\includegraphics[width=0.5\textwidth]{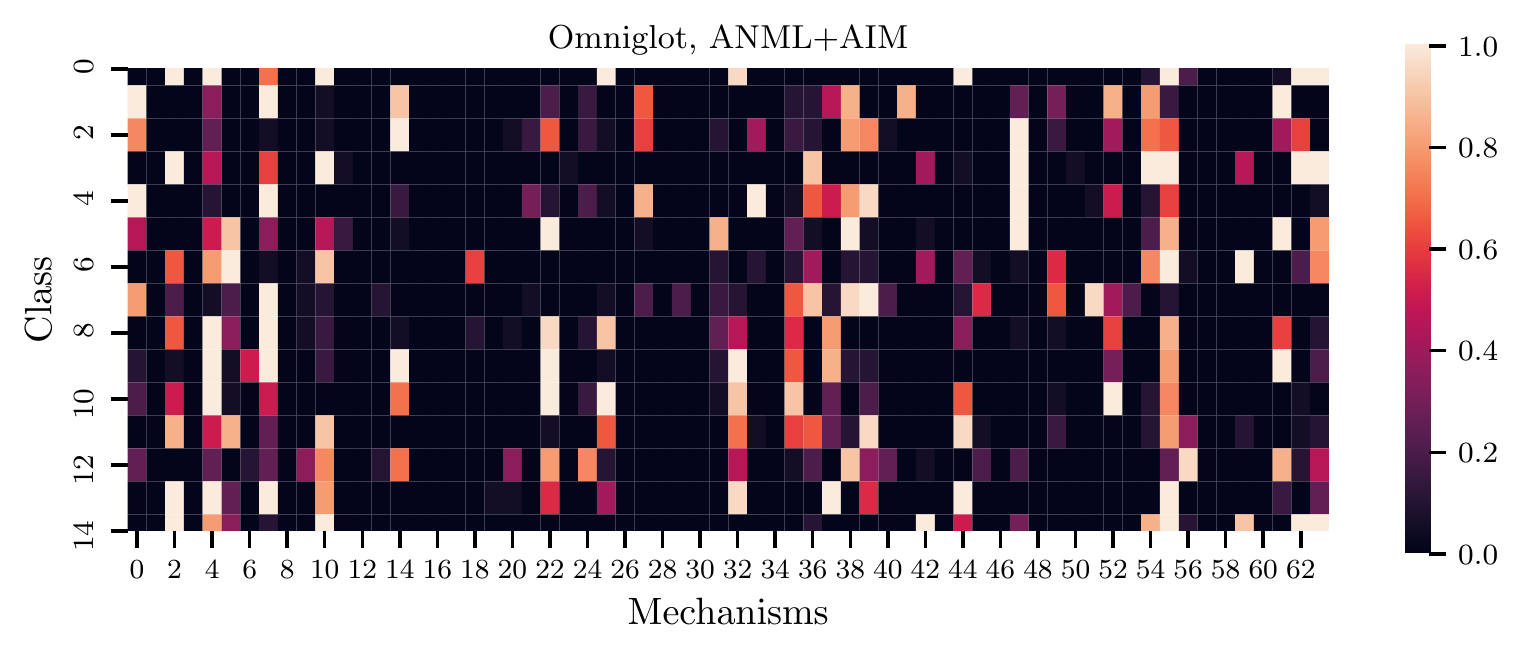}\label{fig:heat_omniglot_anml}}
\caption{Activation of AIM from continual learning. Subset of classes from Omniglot are shown.}
\label{fig:mask-cont_omni}
\end{figure*}

\begin{figure*}[!htbp]
\centering
  	\subfloat[OML+AIM]{\includegraphics[width=0.5\textwidth]{images/heat_cifar100_AIM.pdf}\label{fig:heat_cifar_oml}}
	\hfill
	\subfloat[ANML+AIM]{\includegraphics[width=0.5\textwidth]{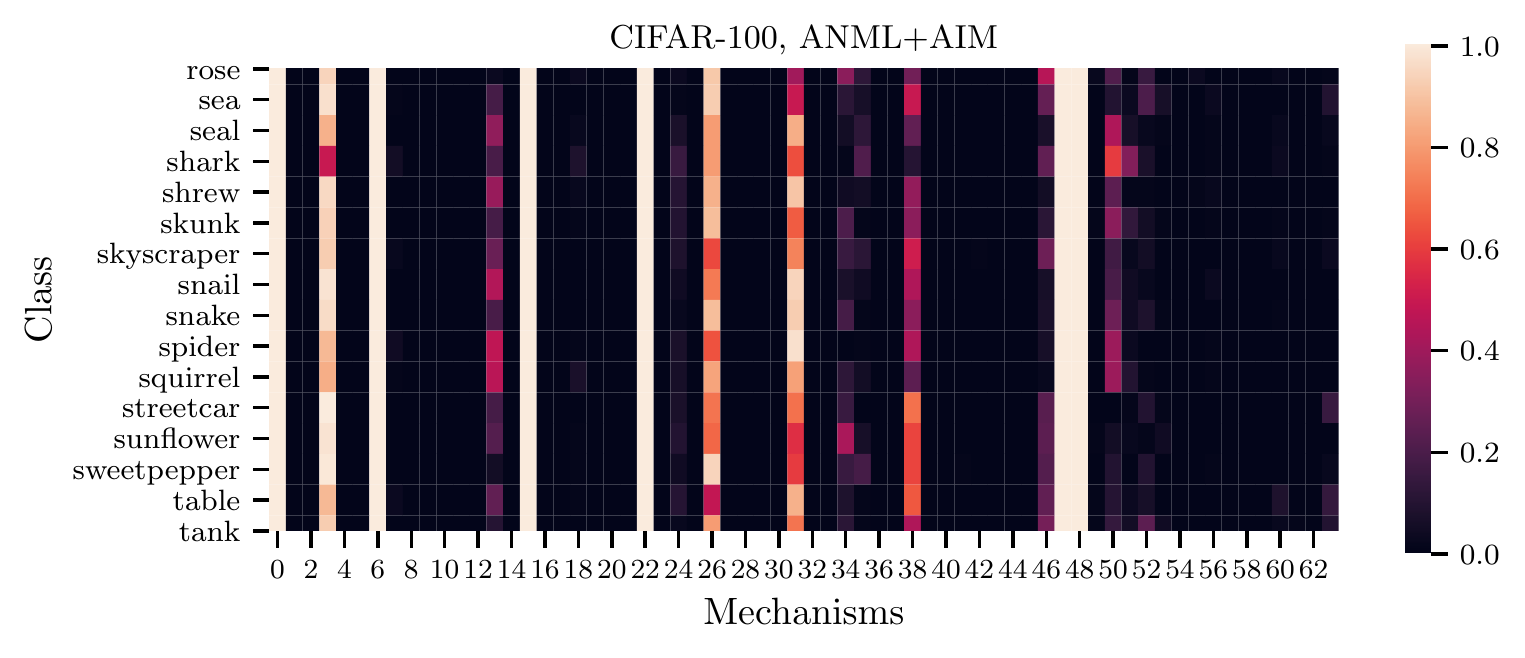}\label{fig:heat_cifar_anml}}
\caption{Activation of AIM from continual learning. Subset of classes from CIFAR-100 are shown.}
\label{fig:mask-cont_cifar}
\end{figure*}

\begin{figure*}[!htbp]
\centering
  	\subfloat[OML+AIM]{\includegraphics[width=0.5\textwidth]{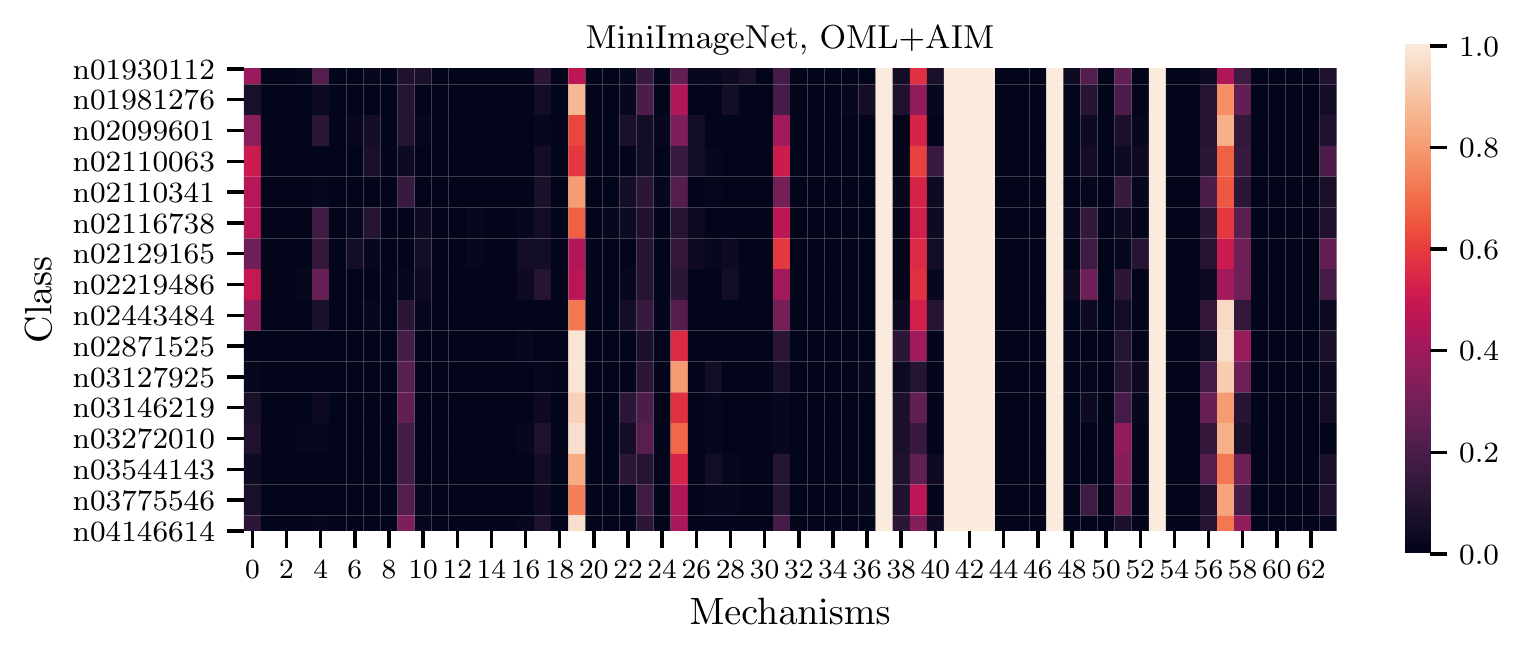}\label{fig:heat_mini_oml}}
	\hfill
	\subfloat[ANML+AIM]{\includegraphics[width=0.5\textwidth]{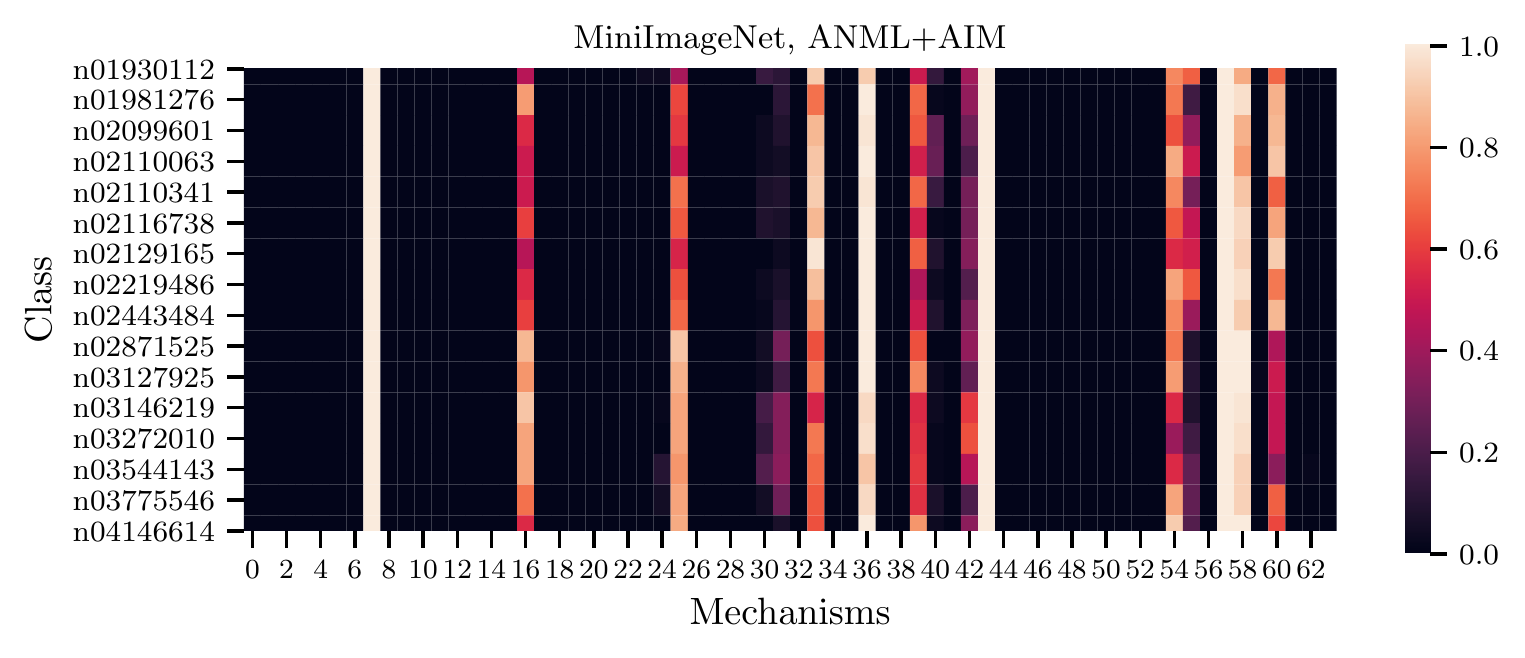}\label{fig:heat_mini_anml}}
\caption{Activation of AIM from continual learning. Subset of classes from MiniImageNet are shown.}
\label{fig:mask-cont_mini}
\end{figure*}

\begin{table*}[!htbp]
\scriptsize
  \caption{Results for varying the stochastic sampling count $K+l$. Zero mean-ed plot is found in the main paper. Throughout the experiment, $K=8$ and $l$ is varied. Average classification accuracies with 95\% confidence intervals on the test-set are shown.}
  \label{tab:sto_count}
  \centering
  \begin{tabular}{cccccc}
    \toprule
    \multirow{2}{*}{Backbone} &  Stochastic sampling count & \multicolumn{2}{c}{MiniImageNet, 5-Way} & \multicolumn{2}{c}{CIFAR-FS, 5-Way}                   \\
       & $K+l$ & 1-shot     & 5-shot & 1-shot     & 5-shot \\
    \midrule
    \multirow{8}{*}{Conv-4-64} 
                & 8  & $61.90\pm0.56\%$  & $74.55\pm0.38\%$ & $70.80\pm0.61\%$ & $80.50\pm0.40\%$    \\
                & 10 & $61.90\pm0.57\%$  & $74.55\pm0.38\%$ & $71.09\pm0.62\%$ & $80.48\pm0.40\%$    \\
                & 12 & $62.13\pm0.58\%$  & $74.66\pm0.38\%$ & $70.68\pm0.62\%$ & $80.36\pm0.39\%$    \\
                & 16 & $61.74\pm0.57\%$  & $74.24\pm0.38\%$ & $69.95\pm0.63\%$ & $79.96\pm0.40\%$    \\
                & 20 & $61.70\pm0.57\%$  & $74.12\pm0.38\%$ & $70.01\pm0.62\%$ & $79.60\pm0.41\%$    \\
                & 24 & $60.73\pm0.57\%$  & $73.58\pm0.39\%$ & $68.82\pm0.63\%$ & $79.21\pm0.40\%$    \\
                & 28 & $60.34\pm0.56\%$  & $73.19\pm0.38\%$ & $67.79\pm0.64\%$ & $78.90\pm0.41\%$    \\
                & 32 & $59.43\pm0.56\%$  & $72.89\pm0.38\%$ & $66.90\pm0.63\%$ & $78.35\pm0.41\%$    \\
    \midrule
    \multirow{8}{*}{WRN-28-10} 
                & 8  & $71.03\pm0.57\%$  & $82.30\pm0.33\%$ & $79.19\pm0.55\%$ & $87.04\pm0.36\%$    \\
                & 10 & $71.22\pm0.57\%$  & $82.25\pm0.34\%$ & $80.20\pm0.55\%$ & $87.34\pm0.36\%$    \\
                & 12 & $71.08\pm0.57\%$  & $82.25\pm0.34\%$ & $80.20\pm0.55\%$ & $87.19\pm0.36\%$    \\
                & 16 & $70.57\pm0.57\%$  & $82.25\pm0.34\%$ & $79.95\pm0.56\%$ & $87.10\pm0.36\%$    \\
                & 20 & $70.38\pm0.57\%$  & $81.86\pm0.34\%$ & $80.17\pm0.56\%$ & $87.07\pm0.36\%$    \\
                & 24 & $70.20\pm0.57\%$  & $81.65\pm0.34\%$ & $80.18\pm0.56\%$ & $86.68\pm0.36\%$    \\
                & 28 & $70.40\pm0.57\%$  & $81.60\pm0.35\%$ & $80.33\pm0.56\%$ & $86.63\pm0.37\%$    \\
                & 32 & $69.34\pm0.55\%$  & $81.19\pm0.34\%$ & $80.13\pm0.56\%$ & $86.22\pm0.38\%$    \\
    \bottomrule
  \end{tabular}
\end{table*}

\begin{table*}[!htbp]
\scriptsize
  \caption{Results for varying the active mechanism count $K$. Zero mean-ed plot is found in the main paper. Throughout the experiment, $K$ is varied and $l=0$. Average classification accuracies with 95\% confidence intervals on the test-set are shown.}
  \label{tab:active_mech}
  \centering
  \begin{tabular}{cccccc}
    \toprule
    \multirow{2}{*}{Backbone} &  Active Mechanism Count & \multicolumn{2}{c}{MiniImageNet, 5-Way} & \multicolumn{2}{c}{CIFAR-FS, 5-Way}                   \\
       & $K$ & 1-shot     & 5-shot & 1-shot     & 5-shot \\
    \midrule
    \multirow{8}{*}{Conv-4-64} 
                & 1  & $25.54\pm0.26\%$  & $62.63\pm0.40\%$ & $36.38\pm0.40\%$ & $68.72\pm0.44\%$    \\
                & 2  & $62.00\pm0.57\%$  & $74.62\pm0.38\%$ & $69.95\pm0.61\%$ & $80.45\pm0.40\%$    \\
                & 4  & $61.93\pm0.56\%$  & $74.54\pm0.38\%$ & $70.30\pm0.60\%$ & $80.37\pm0.39\%$    \\
                & 8  & $61.59\pm0.56\%$  & $74.54\pm0.38\%$ & $70.15\pm0.62\%$ & $80.46\pm0.39\%$    \\
                & 12 & $61.39\pm0.56\%$  & $74.62\pm0.38\%$ & $70.65\pm0.61\%$ & $80.77\pm0.38\%$    \\
                & 16 & $61.60\pm0.55\%$  & $74.67\pm0.38\%$ & $70.66\pm0.60\%$ & $80.68\pm0.39\%$    \\
                & 20 & $61.68\pm0.56\%$  & $74.67\pm0.39\%$ & $70.00\pm0.61\%$ & $80.52\pm0.39\%$    \\
                & 24 & $61.81\pm0.56\%$  & $74.67\pm0.38\%$ & $70.01\pm0.61\%$ & $80.55\pm0.39\%$    \\
                & 28 & $61.81\pm0.56\%$  & $74.63\pm0.39\%$ & $70.74\pm0.60\%$ & $80.46\pm0.39\%$    \\
                & 32 & $61.89\pm0.56\%$  & $74.79\pm0.39\%$ & $70.56\pm0.61\%$ & $80.39\pm0.39\%$    \\
    \midrule
    \multirow{8}{*}{WRN-28-10} 
                & 1  & $61.01\pm0.55\%$  & $73.84\pm0.37\%$ & $72.21\pm0.57\%$ & $81.03\pm0.41\%$    \\
                & 2  & $69.98\pm0.56\%$  & $81.89\pm0.34\%$ & $79.88\pm0.55\%$ & $86.90\pm0.36\%$    \\
                & 4  & $70.03\pm0.56\%$  & $82.10\pm0.33\%$ & $79.53\pm0.55\%$ & $86.25\pm0.37\%$    \\
                & 8  & $69.76\pm0.57\%$  & $82.30\pm0.33\%$ & $79.19\pm0.55\%$ & $86.26\pm0.37\%$    \\
                & 12 & $69.93\pm0.56\%$  & $82.26\pm0.33\%$ & $79.53\pm0.55\%$ & $86.68\pm0.37\%$    \\
                & 16 & $70.24\pm0.55\%$  & $82.15\pm0.33\%$ & $79.19\pm0.55\%$ & $86.74\pm0.37\%$    \\
                & 20 & $69.87\pm0.55\%$  & $82.08\pm0.34\%$ & $79.42\pm0.54\%$ & $86.83\pm0.37\%$    \\
                & 24 & $69.80\pm0.56\%$  & $82.52\pm0.33\%$ & $79.89\pm0.54\%$ & $87.08\pm0.37\%$    \\
                & 28 & $69.58\pm0.55\%$  & $82.30\pm0.33\%$ & $79.66\pm0.54\%$ & $86.88\pm0.37\%$    \\
                & 32 & $70.09\pm0.55\%$  & $82.29\pm0.33\%$ & $80.01\pm0.53\%$ & $87.02\pm0.37\%$    \\
    \bottomrule
  \end{tabular}
\end{table*}

\begin{figure*}[!htbp]
\centering
  	\subfloat{\includegraphics[width=0.45\textwidth]{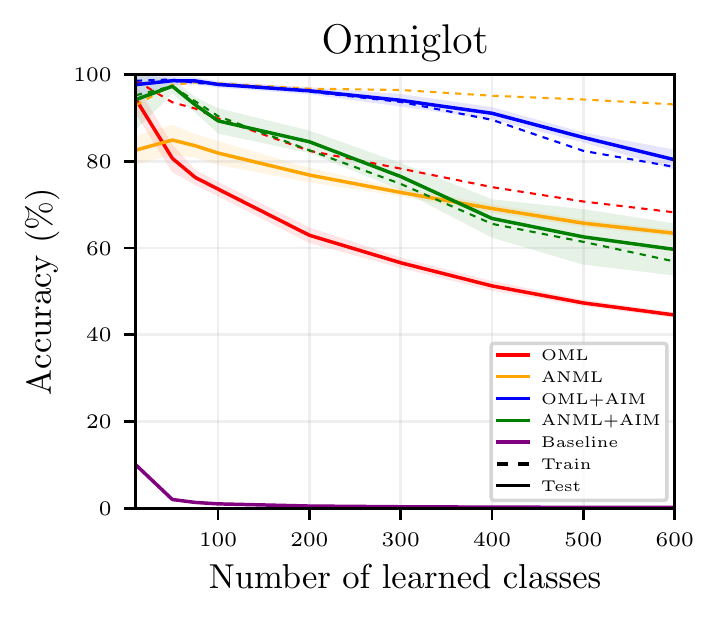}\label{fig:omniglot}}
	\vfill
	\subfloat{\includegraphics[width=0.45\textwidth]{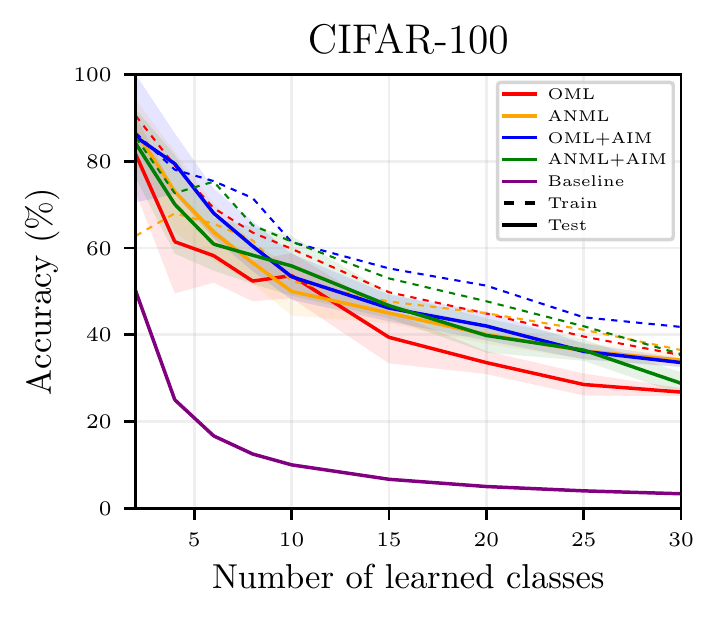}\label{fig:cifar100}}
	\vfill
	\subfloat{\includegraphics[width=0.45\textwidth]{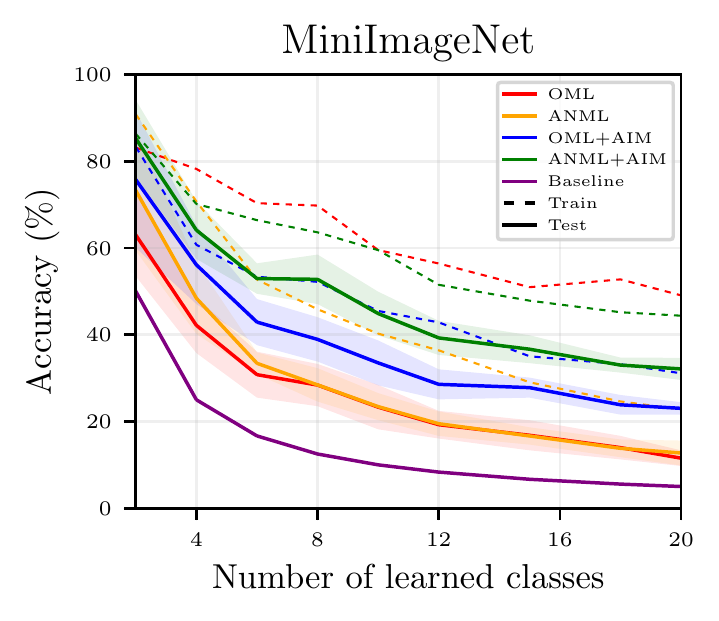}\label{fig:miniimagenet}}
\caption{Evaluation of continual learning methods using dataset of various scales. Meta-test testing (training) trajectories are shown in solid (dashed) lines. All curves are averaged over 10 runs with standard deviation shown.}
\label{fig:benchmark_continual}
\end{figure*}

\begin{table*}[!htbp]
\scriptsize
  \caption{Average meta-testing test accuracy of continual learning on various datasets. During training, trajectory of samples are introduced, i.e.\ meta-test train images are fed to the model sequentially without the usage of rehearsal memory and evaluation using meta-testing test set is performed at the end. Relative increment (decrease) in accuracy through the introduction of AIM is shown in \textcolor{ForestGreen}{green} (\textcolor{red}{red}), e.g.\ when 10 classes in a trajectory is introduced for Omniglot, a relative increment in accuracy of \textcolor{ForestGreen}{+3.45} over $\mathrm{OML}$ is attained when AIM is inserted, shown as $\mathrm{OML+AIM}$. }
  \label{tab:continualz}
  \centering
  \begin{tabular}{cccccccccc}
    \toprule
    \multirow{2}{*}{Method} & \multicolumn{9}{c}{Number of classes} \\
      & 10 & 50 & 75 & 100 & 200 & 300 & 400 & 500 & 600 \\
    \cmidrule{2-10}
      \multicolumn{10}{c}{Dataset: Omniglot} \\
        Baseline & 10.00 & 2.00 & 1.33 & 1.00 & 0.50 & 0.33 & 0.25 & 0.20 & 0.17 \\
        OML      & 94.34 & 80.69 & 76.30 & 73.61 & 62.96 & 56.61 & 51.27 & 47.34 & 44.56 \\
        ANML     & 82.60 & 84.92 & 83.60 & 81.92 & 76.85 & 72.83 & 69.12 & 65.74 & 63.41 \\
        OML+AIM  & 97.70 \textcolor{ForestGreen}{\tiny(+3.45)} & 98.60 \textcolor{ForestGreen}{\tiny(+1.28)} & 98.55 \textcolor{ForestGreen}{\tiny(+5.50)} & 97.75 \textcolor{ForestGreen}{\tiny(+8.41)} & 96.28 \textcolor{ForestGreen}{\tiny(+11.76)} & 94.08 \textcolor{ForestGreen}{\tiny(+17.58)} & 91.09 \textcolor{ForestGreen}{\tiny(+24.26)} & 85.51 \textcolor{ForestGreen}{\tiny(+22.93)} & 80.37 \textcolor{ForestGreen}{\tiny(+20.70)} \\
        ANML+AIM & 94.25 \textcolor{ForestGreen}{\tiny(+11.65)} & 97.32 \textcolor{ForestGreen}{\tiny(+12.40)} & 93.05 \textcolor{ForestGreen}{\tiny(+9.45)} & 89.34 \textcolor{ForestGreen}{\tiny(+7.42)} & 84.52 \textcolor{ForestGreen}{\tiny(+7.67)} & 76.50 \textcolor{ForestGreen}{\tiny(+3.67)} & 66.83 \textcolor{Red}{\tiny(-2.29)} & 62.58 \textcolor{Red}{\tiny(-3.17)} & 59.68 \textcolor{Red}{\tiny(-3.74)} \\
    \midrule
     & 2 & 4 & 6 & 8 & 10 & 15 & 20 & 25 & 30 \\
    \cmidrule{2-10}
      \multicolumn{10}{c}{Dataset: CIFAR-100} \\
        Baseline & 50.00 & 25.00 & 16.67 & 12.50 & 10.00 & 6.67 & 5.00 & 4.00 & 3.33 \\
        OML      & 81.57 & 61.46 & 58.24 & 52.39 & 53.65 & 39.42 & 33.58 & 28.53 & 26.78 \\
        ANML     & 86.86 & 73.03 & 63.66 & 56.60 & 50.01 & 45.02 & 40.06 & 36.29 & 34.15 \\
        OML+AIM  & 85.65 \textcolor{ForestGreen}{\tiny(+4.08)} & 79.46 \textcolor{ForestGreen}{\tiny(+18.00)} & 68.03 \textcolor{ForestGreen}{\tiny(+9.79)} & 60.44 \textcolor{ForestGreen}{\tiny(+8.04)} & 53.39 \textcolor{Red}{\tiny(-0.26)} & 46.16 \textcolor{ForestGreen}{\tiny(+6.74)} & 42.02 \textcolor{ForestGreen}{\tiny(+8.43)} & 36.17 \textcolor{ForestGreen}{\tiny(+7.64)} & 33.59 \textcolor{ForestGreen}{\tiny(+6.81)} \\
        ANML+AIM & 84.10 \textcolor{Red}{\tiny(-2.76)} & 70.10 \textcolor{Red}{\tiny(-2.93)} & 60.90 \textcolor{Red}{\tiny(-2.76)} & 58.35 \textcolor{ForestGreen}{\tiny(+1.76)} & 55.88 \textcolor{ForestGreen}{\tiny(+5.87)} & 46.72 \textcolor{ForestGreen}{\tiny(+1.70)} & 39.84 \textcolor{Red}{\tiny(-0.23)} & 36.47 \textcolor{ForestGreen}{\tiny(+0.18)} & 28.81 \textcolor{Red}{\tiny(-5.34)} \\
    \midrule
    & 2 & 4 & 6 & 8 & 10 & 12 & 15 & 18 & 20 \\
    \cmidrule{2-10}
      \multicolumn{10}{c}{Dataset: MiniImageNet} \\
        Baseline & 50.00 & 25.00 & 16.67 & 12.50 & 10.00 & 8.33 & 6.67 & 5.56 & 5.00 \\
        OML      & 63.00 & 42.17 & 30.80 & 28.40 & 23.31 & 19.26 & 16.82 & 13.97 & 11.54 \\
        ANML     & 73.25 & 48.42 & 33.42 & 28.44 & 23.43 & 19.49 & 16.66 & 13.81 & 12.73 \\
        OML+AIM  & 75.75 \textcolor{ForestGreen}{\tiny(+12.75)} & 56.13 \textcolor{ForestGreen}{\tiny(+13.96)} & 42.92 \textcolor{ForestGreen}{\tiny(+12.12)} & 38.94 \textcolor{ForestGreen}{\tiny(+10.53)} & 33.52 \textcolor{ForestGreen}{\tiny(+10.20)} & 28.57 \textcolor{ForestGreen}{\tiny(+9.30)} & 27.81 \textcolor{ForestGreen}{\tiny(+10.99)} & 23.85 \textcolor{ForestGreen}{\tiny(+9.89)} & 23.03 \textcolor{ForestGreen}{\tiny(+11.48)} \\
        ANML+AIM & 85.25 \textcolor{ForestGreen}{\tiny(+12.00)} & 64.13 \textcolor{ForestGreen}{\tiny(+15.71)} & 52.97 \textcolor{ForestGreen}{\tiny(+19.56)} & 52.79 \textcolor{ForestGreen}{\tiny(+24.35)} & 44.90 \textcolor{ForestGreen}{\tiny(+21.47)} & 39.28 \textcolor{ForestGreen}{\tiny(+19.79)} & 36.67 \textcolor{ForestGreen}{\tiny(+20.01)} & 33.01 \textcolor{ForestGreen}{\tiny(+19.20)} & 32.13 \textcolor{ForestGreen}{\tiny(+19.40)} \\
\bottomrule
  \end{tabular}
\end{table*}

\end{document}